\documentclass{article} 
\usepackage{iclr2027_conference,times}

\usepackage{amsmath,amsfonts,bm}

\def\eqref#1{equation~\ref{#1}}

\def\1{\bm{1}}

\DeclareMathAlphabet{\mathsfit}{\encodingdefault}{\sfdefault}{m}{sl}
\SetMathAlphabet{\mathsfit}{bold}{\encodingdefault}{\sfdefault}{bx}{n}

\usepackage{hyperref}
\usepackage{url}

\usepackage[utf8]{inputenc} 
\usepackage[T1]{fontenc}    
\usepackage{hyperref}       
\usepackage{url}            
\usepackage{booktabs}       
\usepackage{amsfonts}       
\usepackage{nicefrac}       
\usepackage{microtype}      
\usepackage{xcolor}         

\usepackage[table]{xcolor}
\usepackage{subcaption}
\usepackage{booktabs, makecell, multirow, graphicx, caption, subcaption}
\usepackage{rotating}        
\usepackage{amsmath}
\usepackage[most]{tcolorbox}
\tcbuselibrary{skins}
\newtcolorbox{findingbox}[1]{%
  enhanced, rounded corners,
  colback=blue!4!white, colframe=blue!45!black,
  leftrule=2pt, rightrule=0.2pt, toprule=0.2pt, bottomrule=0.2pt,
  left=3pt, right=3pt, top=1pt, bottom=1pt,
  fonttitle=\bfseries\small, title={Finding #1},
  before skip=4pt, after skip=4pt,
}

\definecolor{humanrow}{RGB}{255, 243, 205}   
\definecolor{sectionbg}{RGB}{240, 244, 248}  
\definecolor{toprule}{RGB}{30,  100, 180}    
\definecolor{bestcol}{RGB}{210, 232, 210}    

\usepackage{hyperref}
\usepackage{xcolor}
\usepackage{fontawesome5} 

\hypersetup{
    colorlinks=true,
    linkcolor=black,
    citecolor=black,
    urlcolor=blue!70!black
}

\def\DatasetName{ActionLens}
\title{\DatasetName: Diagnosing Spatial-Temporal Binding Failures in Vision-Language Models}

\author{%
  \parbox{\linewidth}{\centering
  \vspace*{1.5em}%
    \textbf{Gueter Josmy Faure}$^{1}$ \quad
    \textbf{Min-Hung Chen}$^{2}$ \quad
    \textbf{Hao Ping Wang}$^{1}$ \\
    \textbf{Timoth{\'e}e Lardy}$^{1}$ \quad
    \textbf{Hung-Ting Su}$^{1}$ \quad
    \textbf{Winston H. Hsu}$^{1}$ \\[1ex]
    \textnormal{$^{1}$National Taiwan University \qquad $^{2}$NVIDIA} \\[1.8ex]
    \textnormal{\small
      \href{https://joslefaure.github.io/actionlens}{\faGlobe\ \textsf{Project Page}} \qquad
      \href{https://github.com/joslefaure/lmms-eval/tree/add-actionlens-dataset}{\faGithub\ \textsf{Code}}
\qquad
      \href{https://huggingface.co/datasets/anonymous-dataset-submission/actionlens}{\faDatabase\
\textsf{Dataset}}
    }
  }
}

\iclrfinalcopy 
\begin{document}

\maketitle

\begin{figure*}[htbp]
    \centering
    \includegraphics[width=.8\linewidth]{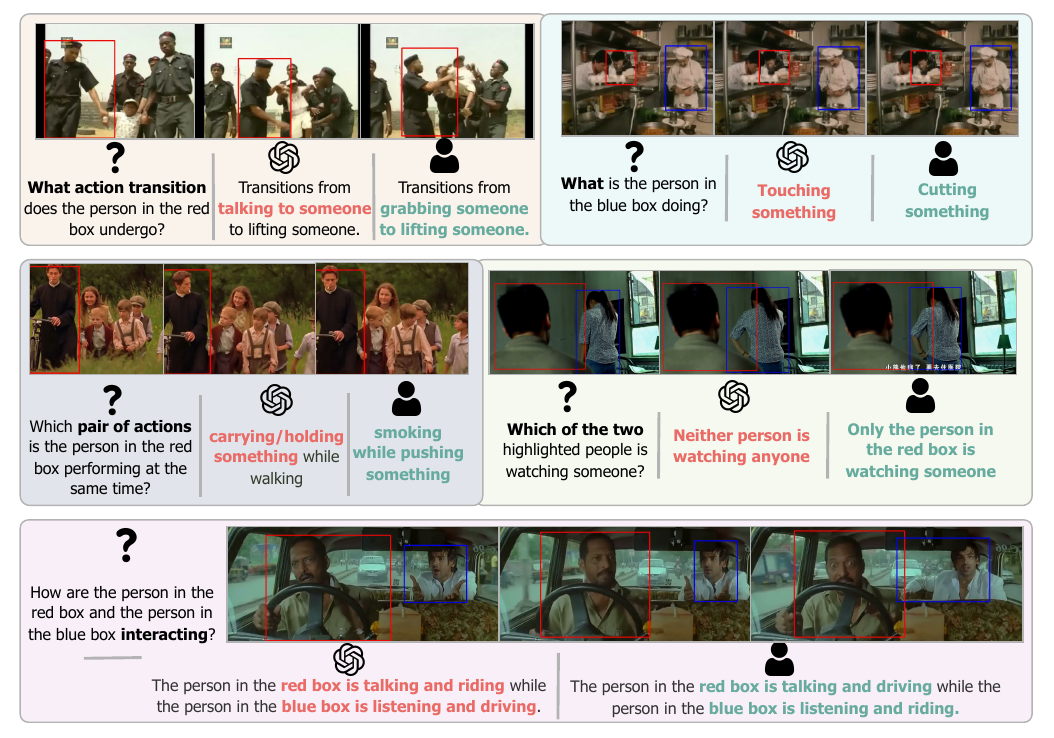}
    \caption{\textbf{\DatasetName{} asks questions humans answer in seconds.}
    One example per diagnostic. \textcolor{red}{Red}: VLM answer. \textcolor{teal}{Teal}: human answer. (These are MCQs, not free-form text.)}
    \label{fig:placeholder}
\end{figure*}

\begin{abstract}
Video-capable vision-language models score above 80\% on popular benchmarks yet struggle with \emph{spatial-temporal binding}: associating the right action with the right person at the right moment.
We introduce \DatasetName{}, a diagnostic benchmark of 6,701 multiple-choice video questions spanning five targeted diagnostics: transition detection, actor-specific identification, concurrent action binding, directed interaction reasoning, and gaze detection.
Ground-truth answers are derived deterministically from 1.58 million per-second, per-person annotations.
Fourteen rounds of human quality engineering raised answer clarity from 53\% to above 90\% human accuracy.
Across 20 VLMs, the full-set leader scores 68.8\%; on the human-reviewed subset, it scores 65.9\% versus 91.0\% for the pooled human reference. Gaze detection remains near chance against 89.6\% human accuracy.
On actor disambiguation, reference-interface controls show that relational descriptions recover 5.55--13.25 points over static coordinates, confirming a substantial numeric-parsing penalty; yet visual boxes still lead every model by 1.15--6.50 points, exposing a residual unboxed actor-resolution gap.
A binding-trap analysis shows models systematically select the wrong actor's action.
\DatasetName{} provides diagnostic measurements of these distinct failure modes across model families and scales for direct comparison.
We release all data, code, and evaluation scripts at \url{https://anonymous.4open.science/r/lmms-eval-2276}, with \DatasetName{} integrated into \texttt{lmms-eval}~\cite{zhang2024lmmseval}.
\end{abstract}

\section{Introduction}
\label{sec:introduction}

Watch someone pick up a phone.  In the span of seconds, you register
\emph{who} grabbed it (not their neighbour), \emph{what} changed (their hand
was empty a moment ago), and whether they are calling someone or just
scrolling, all without conscious effort. This is
\emph{spatial-temporal binding}: the association of the right action
with the right agent at the right moment. It is so automatic in humans that
we rarely think of it as a skill. It's precisely the skill that current
video-language models lack.

State-of-the-art VLMs score impressively on popular video benchmarks, often above 80\% on VideoMME~\cite{fu2024videomme} and
MVBench~\cite{li2024mvbench}, yet those benchmarks measure something
closer to \emph{scene-level recognition}: identifying what is broadly
happening, which object appears...  They do not ask
a model to watch two people and answer: \emph{``What is the person [attribute] doing?''}, a question any child answers in a glance and one that
exposes a different set of model competencies.

We introduce \textbf{\DatasetName{}}, a diagnostic benchmark designed to
make exactly this distinction explicit.
\DatasetName{} decomposes spatial-temporal binding into five targeted
diagnostics:
\textbf{D1}~(Transition Sensitivity) tests whether a model notices that an
action \emph{changed}, not just what the majority action is;
\textbf{D2}~(Actor Disambiguation) asks which of two co-present people is
performing a given action, with the \emph{other} person's action as a
distractor;
\textbf{D3}~(Concurrent Binding) probes whether a model can report
\emph{both} actions of a person who is multitasking;
\textbf{D4}~(Interaction Reasoning) requires resolving the asymmetric roles
in a directed two-person exchange;
\textbf{D5}~(Gaze Detection) asks which highlighted person is watching
someone else. Every ground-truth answer is derived \emph{deterministically} from AVA
v2.2~\cite{gu2018ava}, a corpus of 1.58 million per-second, per-person
action annotations.  No language model generated our labels.  A 14-round
human quality-engineering process raised question clarity from 53\% to
above 90\% human accuracy, and each diagnostic is defended against known
shortcuts: binding traps, physically compatible
distractor pairs, and positional bias.  The result is 6,701 questions
across 6,701 video clips that cannot be passed by exploiting visual scene biases or other shortcuts.

We evaluate 20 VLMs spanning 4B to 38B parameters alongside GPT-5.2 and
Gemini~3~Flash.
The strongest open-weight model reaches 68.8\% on the full set; on the
human-reviewed subset, its 65.9\% trails the 91.0\% pooled human reference by 25 points.
The gap is not uniform: models score near 70\% on temporal transitions and
concurrent binding, but collapse on actor disambiguation (as low as 34\%)
and gaze detection (near random for most models, despite the pooled human
reference scoring 89.6\%). Reference-interface controls compare tracked visual
boxes, static midpoint coordinates, and ordinary relational descriptions.
Two robust conclusions emerge for actor disambiguation. First, relational descriptions recover
5.55--13.25 points over static coordinates, confirming that numeric
parsing causes a substantial avoidable penalty. Second, every model remains
1.15--6.50 points below visual boxes, exposing a residual unboxed
actor-resolution gap. Interaction reasoning is model-dependent, showing that two-person binding
remains highly sensitive to how the actors are specified.

\DatasetName{} is a diagnostic probe: each score measures a targeted
operation. A model that scores 65\% overall but 34\%
on actor disambiguation has a specific, actionable failure mode. Naming the failure
is the first step toward fixing it.

\paragraph{Contributions.}
We contribute \textbf{(1)}~\textbf{\DatasetName{}, a diagnostic benchmark} of 6,701
multiple-choice video questions spanning five spatial-temporal binding capabilities,
with deterministic ground truth and a documented 14-round quality-engineering process;
\textbf{(2)}~\textbf{a systematic evaluation of 20 VLMs}, revealing that
spatial-binding diagnostics and gaze detection remain far below human performance
regardless of model scale or family;
\textbf{(3)}~\textbf{reference-interface controls} establishing a substantial
numeric-parsing penalty on actor disambiguation while showing that explicit visual anchors still
outperform relational descriptions for every tested model on that diagnostic;
\textbf{(4)}~\textbf{extensive ablations}: a frame-budget sweep showing which
diagnostics require temporal coverage versus a single frame, a text-only audit of
vision-dependency, and a binding-trap analysis showing most models systematically
select the wrong actor's action; and
\textbf{(5)}~\textbf{a one-command reproduction pipeline} integrated into
\texttt{lmms-eval}~\cite{zhang2024lmmseval}, allowing any researcher to evaluate any
supported model on the \DatasetName{} suite with a single script.

\section{Related Work}
\label{sec:related}

\textbf{Video question answering benchmarks.}
The dominant paradigm for evaluating video understanding models is multiple-choice
question answering.  Early benchmarks such as MSVD-QA and MSRVTT-QA~\cite{xu2017video}
convert open-ended video captions into QA pairs but rely on lexical matching metrics
that reward action-name recall over genuine understanding.
ActivityNet-QA~\cite{yu2019activitynet} scales this to 58,000
question–answer pairs drawn from ActivityNet~\cite{caba2015activitynet}, yet answers
are single words evaluated with exact match, obscuring how well a model \emph{reasons}.  NExT-QA~\cite{xiao2021nextqa} introduces causal and temporal
question types with five-option MCQ, revealing that models rely on appearance
statistics rather than temporal reasoning.  STAR~\cite{wu2021star} isolates four
reasoning types (interaction, sequence, prediction, feasibility) using programmatic
question generation from situated captions, prefiguring ActionLens's construction
philosophy.  MVBench~\cite{li2024mvbench} broadens coverage to 20 tasks generated
by a static-to-dynamic conversion of existing annotations, while VideoMME~\cite{fu2024videomme}
provides multi-duration evaluation (11 s to 1 hour) with expert-annotated questions
across 30 sub-domains. Perception Test~\cite{patraucean2023perceptiontest}
combines video QA with tracking and temporal action annotations.
EgoSchema~\cite{mangalam2023egoschema} stresses \emph{temporal
certificate length}: its QA pairs require attending to 3-minute ego-centric clips.
TempCompass~\cite{liu2024tempcompass} targets temporal perception,
constructing \emph{conflicting videos} that share static content but differ in speed
or direction to neutralise single-frame shortcuts.

Despite this breadth, \emph{no existing benchmark jointly requires per-person spatial
grounding and temporal transition detection with programmatically verified ground
truth}.  Benchmarks that address multiple people (e.g., MVBench's action counting task)
still treat the scene holistically, not binding actions to specific individuals.
ActionLens directly operationalises this missing capability through five diagnostics
built on per-second, per-person annotations.

\textbf{Dense video annotation.}
AVA~\cite{gu2018ava} provides 80 atomic visual actions annotated per-second and per-person bounding box across 299 Hollywood films, 1.58 million action instance annotations in total, the densest public source of person-level action labels.
Our benchmark derives every ground-truth answer deterministically from these annotations, inheriting their reliability and avoiding the LLM-label noise that plagues many recent benchmarks~\cite{li2024mvbench,fu2024videomme}.

\textbf{Spatial grounding and compositional reasoning.}
A growing body of work probes whether vision-language models can \emph{bind}
properties to the correct entities, not merely recognise them.  Winoground~\cite{thrush2022winoground}
shows that image–text models systematically fail to distinguish ``a dog chasing a
cat'' from ``a cat chasing a dog'' despite recognising the individual objects,
directly motivating our actor–role confusion hypothesis.  VSR~\cite{liu2023vsrbenchmark}
benchmarks binary visual spatial relationships (above, below, left of, etc.) in
static images and finds frontier models at best $\sim$70\%, illustrating the difficulty of spatial attribute binding even without
temporal dynamics.  SpatialBench~\cite{cai2024spatialbot} extends spatial grounding
to 3D scenes; we extend the same question — \emph{can models associate spatial
annotations with the right entity?} — to the video domain through controlled
comparisons of tracked visual boxes, static midpoint coordinates, and relational
descriptions.

\textbf{Vision-language models for video.}
LLaVA-Video~\cite{zhang2024llavavideo}, Qwen3.5~\cite{qwen2026qwen35}, InternVL3~\cite{chen2025internvl3},
Gemini~3~\cite{team2026gemini3flash}, and the  GPT series~\cite{openai2024gpt4o} represent the
current frontier in video-language modelling, combining large language models with
video encoders capable of processing dozens of frames.
Despite strong global performance, these models are evaluated on benchmarks that reward scene-level recognition rather than per-person spatial binding — the gap \DatasetName{} is designed to expose.

\begin{table*}[t]
\centering
\small
\caption{%
  \textbf{ActionLens is the only video benchmark with per-person spatial grounding,
  programmatic ground truth, and iterated human validation.}
  \checkmark\ =~yes; $\circ$~=~partial; --~=~no.%
}
\label{tab:benchmark_comparison}
\begin{tabular}{l r r c c c c}
\toprule
\textbf{Benchmark} &
\makecell[c]{\textbf{QA}\\\textbf{pairs}} &
\makecell[c]{\textbf{Clips}} &
\makecell[c]{\textbf{Per-sec.}\\\textbf{temporal}\textbf{GT}} &
\makecell[c]{\textbf{Per-person}\\\textbf{spatial}\textbf{GT}} &
\makecell[c]{\textbf{Multi-person}\\\textbf{compositional}} &
\makecell[c]{\textbf{Iterated}\\\textbf{human}\\\textbf{validation}} \\
\midrule
VideoMME       & 2,700  & 900   & --        & --        & --           & -- \\
MVBench        & 4,000  & --    & --        & --        & $\circ$     & -- \\
EgoSchema      & 5,031  & 5,031 & --        & --        & --          & -- \\
STAR           & 60,000 & 22,000& $\circ$   & --        & --          & -- \\
PerceptionTest & 11,619 & 691 & \checkmark & -- & $\circ$  & -- \\
TempCompass    & 7,540  & 500   & \checkmark& --        & --      & -- \\
\midrule
\textbf{ActionLens (ours)} &
\textbf{6,701} &
\textbf{6,701} &
\checkmark & \checkmark & \checkmark & \checkmark \\
\bottomrule
\end{tabular}
\end{table*}

\section{The \DatasetName{} Benchmark}
\label{sec:benchmark}

\subsection{Design Philosophy}
\label{subsec:philosophy}

Fine-grained video understanding is not a single skill, it is a
\emph{bundle} of distinct capabilities that existing benchmarks aggregate
into a single score, obscuring which capabilities are actually lacking.
\DatasetName{} is built on the premise that progress requires
\emph{diagnostic decomposition}: each diagnostic should isolate exactly one
spatial-temporal binding operation, admit an unambiguous correct answer, and
be hard enough to discriminate strong models.
The five diagnostics introduced in Section~\ref{sec:introduction} each
test one such operation; the following subsections specify their construction.

Every ground-truth answer is derived \emph{deterministically} from AVA
v2.2~\cite{gu2018ava}, a corpus of 1.58 million per-second, per-person
action annotations across 299 Hollywood films.  Since the ground truth is
a pure function of the annotations, no human labelling is required
for answer generation.
This distinguishes \DatasetName{} from recent work that relies on LLM-generated
labels~\cite{li2024mvbench,fu2024videomme}, whose noise is difficult to audit.

\subsection{The Five Diagnostics}
\label{subsec:diagnostics}

\textbf{D1: Transition Sensitivity (194 questions).}
A model that classifies actions frame-by-frame may correctly recognise each
individual pose yet miss that the pose \emph{changes} mid-clip and
reports the majority action. This diagnostic directly tests this blind spot.
Each question presents a variable-length clip (4–10\,s, mean 7.1\,s) with a
red bounding box tracking one person.  The question asks what action
transition the person undergoes.  The options always include:
(i)~the correct before→after sequence, (ii)~the \emph{temporal reversal}
after→before, and (iii–iv)~two \emph{static traps} claiming the person
performs only one of the two actions throughout.  A model that ignores
temporal order cannot distinguish the first two options; a model that
averages over frames will be pulled toward the static traps.

Transition boundaries are located by a \emph{locally-stable boundary scan}:
a boundary qualifies only when the old action is present (new absent) for at
least two consecutive seconds before it, and the new action is present
(old absent) for two seconds after.  The segment is then centred on the boundary and extended to the
limits of the stable regions, yielding natural variable-length clips.
Transitions involving exclusively auditory actions (talk, listen) are
excluded since they can't be resolved in muted video.  The retained
items span posture changes (74\%; e.g., \emph{stand→bend/bow}), object-interaction changes (21\%; e.g.,
\emph{touch→carry/hold}), locomotion changes (2\%) and others (3\%).

\textbf{D2: Actor-Specific Action Disambiguation (2,000 questions).}
Recognising that a scene contains `carrying' and `watching' is easier than
knowing \emph{which} person is carrying and which is watching. This diagnostic isolates the binding step. Each clip contains two people, each tracked by a differently-coloured
bounding box (red and blue).  The question asks directly: \emph{``What is
the person in the blue box doing?''}  Crucially, the red person's action
always appears as a distractor option, creating a \emph{binding trap}: a
model that recognises the action but fails to bind it to the correct person
will be drawn toward this distractor.  All distractors are verified to be
false for the blue person at the query second, so only genuine person-action
binding resolves the question.

\textbf{D3: Concurrent Action Binding (2,000 questions).}
People routinely \emph{multitask}: walk while carrying a bag, sit while using
a phone. This diagnostic asks whether models can recognise both actions of a person who
is simultaneously performing exactly two visual actions, not just the most
salient one.

The correct option is an \emph{action pair}; the three distractors each
change one or both actions.  Distractor selection enforces two constraints:
(a)~neither distractor action is a true action of the highlighted person, and
(b)~distractor pairs are \emph{physically compatible}. Impossible co-occurrence
combinations such as `standing while lying down' or `walking while crouching'
are rejected. This prevents
models from using biomechanical impossibility as a shortcut rather than
watching the video.

\textbf{D4: Directed Interaction Reasoning (1,611 questions).}
Each clip pairs two highlighted people with opposite AVA communication labels
(talking and listening) and distinct visible co-actions (e.g., throwing and
watching). The question asks how both people are interacting; its choices
vary which person is assigned each role and co-action.

Four distractor types are used: role swap, co-action swap, scene alternative, and a 3-option variant when the scene pool is too small.


\textbf{D5: Gaze Detection (896 questions).}
Gaze is a social signal that models have proven difficulty interpreting.
This diagnostic presents two highlighted people and asks which of them is watching someone.
The four options cover all binary combinations: only red, only blue, both,
or neither. Ground truth uses AVA's per-person binary gaze label with a
3-second temporal consistency filter.
\subsection{Quality Engineering}
\label{subsec:quality}
A benchmark is only as reliable as its construction process.  We conducted
\textbf{14 iterative human-review rounds}, each targeted at a specific failure
mode revealed by the previous round.  Table~\ref{tab:fix_rounds} (in Appendix) summarises
the progression.  We document this process as a methodological contribution:
iterative refinement with explicit failure-mode tracking is a reliable way to
build a diagnostic benchmark that is simultaneously solvable by humans and
discriminating for models.

\begin{figure*}[t]
  \centering
  \begin{subfigure}[t]{0.35\linewidth}
    \centering
    \includegraphics[width=\linewidth]{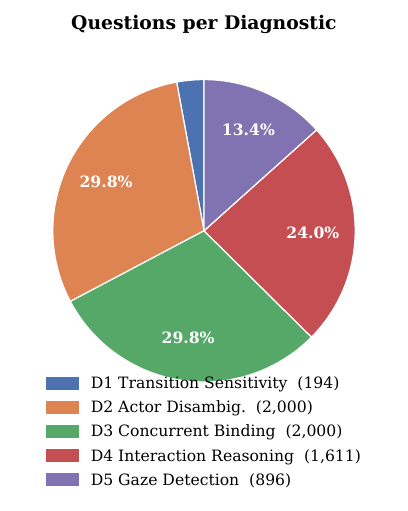}
    \caption{%
      How many questions per diagnostic.
      Actor Disambiguation and Concurrent Binding dominate ($N\!=\!2{,}000$ each);
      Transition Sensitivity is the smallest.
    }
    \label{fig:question_distribution}
  \end{subfigure}
  \hfill
  \begin{subfigure}[t]{0.63\linewidth}
    \centering
    \includegraphics[width=\linewidth]{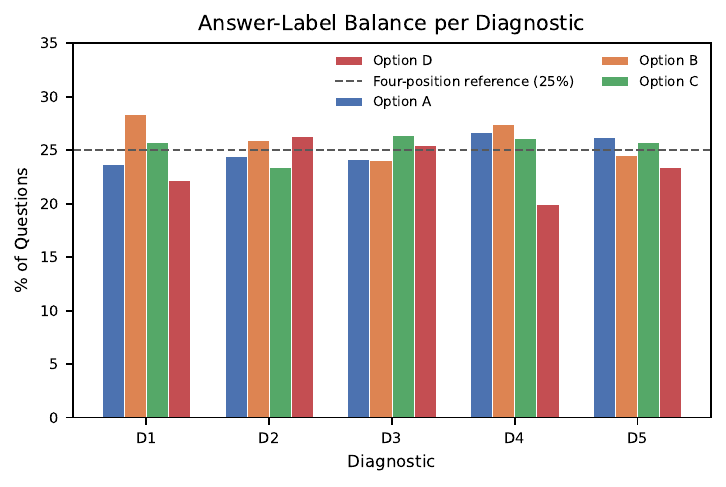}
    \caption{%
      Answer-label balance. All diagnostics stay within $\pm5.1$\% of the
      25\% four-position reference, with no answer-position shortcut to exploit.
      The slight option-D dip for Interaction Reasoning is structural: 21\% of its items have
      only three options.
    }
    \label{fig:answer_balance}
  \end{subfigure}
  \caption{%
    \textbf{\DatasetName{} dataset composition.}
    Actor Disambiguation and Concurrent Binding carry the most weight; answer distributions are near-uniform
    across A--D for every diagnostic.
  }
  \label{fig:dataset_composition}
  \vspace{-4pt}
\end{figure*}

The starting point was a 281-sample pilot review with overall human accuracy
of~53\%.  Rounds~1--6 targeted
the most severe structural failures (degenerate labels, distractor
contamination, auditory leakage) and raised accuracy to 74\%.  Rounds~7--14
addressed subtler issues: tracker identity drift, physically-confusable
action pairs, and temporal label noise.  The refinement also \emph{pruned}
two proposed diagnostics: \textbf{Temporal Localisation}, whose action-onset
answers lacked a defensible tolerance window, and the standalone
\textbf{Talker-Listener Detection} task, whose role-only answers depended on
speech cues that muted video cannot reliably establish. Interaction Reasoning
(D4) retains talk/listen
labels within composite choices that also require binding distinct visible
co-actions to the two people.

Answer-label distributions are audited across all tasks: the most
over-represented option deviates from the 25\% four-position baseline by at most
5.1\% for Interaction Reasoning, and at most 3.4\% for all other tasks
(Fig.~\ref{fig:answer_balance}). Its slight imbalance is structural: the
21\% 3-option subset leaves option D under-represented by design.

\subsection{Dataset Statistics}
\label{subsec:stats}

Figure~\ref{fig:dataset_composition} reports the per-diagnostic statistics.
\DatasetName{} contains \textbf{6,701 unique questions} drawn from \textbf{64 unique videos}. 
All main-task clips are 5 seconds, except Transition Sensitivity (4–10\,s, variable). The benchmark spans
73 unique actions, with Concurrent Binding correct answers covering 86 distinct
co-action pairs.  Every question is paired with a boxed video segment and, in the static-coordinate reference condition, with the same
unboxed segment and textual bounding-box coordinates. The pooled-human-reference subset totals \textbf{1,194 items}: all 194 Transition Sensitivity items plus
250 randomly selected items per remaining diagnostic (seeded 
video-diverse round-robin sampler).

\section{Experiments \& Ablations}
\paragraph{Experimental setup.}
We evaluate \textbf{20 VLMs} spanning three size tiers:
large open-source ($\geq$27B: InternVL3.5-38B, Qwen3.5-27B, Qwen3-VL-32B,
Qwen2.5-VL-32B, Gemma-4-31B, Gemma-3-27B, LLaVA-Video-32B),
mid-size (7--9B: Qwen3.5-9B, Qwen3-VL-8B, InternVL3.5-8B, VideoLLaMA3-7B,
LLaVA-OneVision-7B, LLaVA-Video-7B), and
small ($\leq$4B: Qwen3.5-4B, Qwen3-VL-4B, InternVL3.5-4B, Gemma-4-E4B,
Gemma-3-4B), plus GPT-5.2~\cite{openai2026gpt5} and
Gemini~3~Flash~\cite{team2026gemini3flash} evaluated on the 1,194-item
human-review subset.
All open-source models are run via \texttt{lmms-eval}~\cite{zhang2024lmmseval}
with a uniform prompt template, 32 uniformly-sampled frames, and \texttt{bfloat16}
precision on one NVIDIA H100 GPU (two for $\geq$38B models).
The metric is \textbf{exact-match accuracy}; Full setup details, prompt templates, and per-run configurations are provided
in Appendix~\ref{app:setup}.
The review subset preserves the broad model ordering, although its
question-weighted scores can shift by several points
(Appendix~\ref{app:subset_scope}).

\begin{table*}[t]
\centering
\small
\setlength{\tabcolsep}{5pt}
\caption{%
  \textbf{\DatasetName{} results.}
  Per-diagnostic and question-micro accuracy (\%) for all 20 VLMs, grouped
  by size. Avg. = correct answers divided by questions in the relevant evaluation set
  (6{,}701 for full-set models; 1{,}194 for the pooled human reference and review-subset models).
  Interaction Reasoning (D4) has 26.7\% random accuracy because 334
  of its 1,611 questions have three options; the full-set weighted random
  accuracy is 25.4\%. \textbf{Bold}: best full-dataset result per
  column; \underline{underline}: best within group. Model sources and
  configurations are listed in Appendix~\ref{app:setup}.%
}
\label{tab:main_results}
\begin{tabular}{l r r r r r r}

\specialrule{1.5pt}{0pt}{0pt}  

\rowcolor{white}
\textbf{Model} &
\makecell[c]{\textbf{Transition}\\\textbf{Sensitivity}} &
\makecell[c]{\textbf{Actor}\\\textbf{Disambig.}} &
\makecell[c]{\textbf{Concurrent}\\\textbf{Binding}} &
\makecell[c]{\textbf{Interaction}\\\textbf{Reasoning}} &
\makecell[c]{\textbf{Gaze}\\\textbf{Detection}} &
\makecell[c]{\textbf{Avg.}} \\

\midrule

\rowcolor{sectionbg}
\multicolumn{7}{l}{\textit{\small Review subset (1,194 items)}} \\[1pt]

\rowcolor{humanrow}
\textbf{Pooled human reference} $\boldsymbol{\star}$
  & \textbf{84.5} & \textbf{92.4} & \textbf{94.0} & \textbf{92.8} & \textbf{89.6}
  & \textbf{91.0} \\

GPT-5.2          & 68.8 & 76.0 & 76.4 & 70.0 & 47.2 & 67.6 \\
Gemini 3 Flash & 58.3 & 54.4 & 52.4 & 27.6 & 33.2 & 44.6 \\
Gemma-4-31B       & 63.9 & 66.0 & 74.4 & 61.6 & 46.0 & 62.3 \\
LLava-Video-32B & 60.3 & 47.2 & 40.8 & 47.6 & 28.4 & 44.1 \\
InternVL3.5-38B  & 68.0 & 72.8 & 67.6 & 68.4 & 31.2 & 61.3 \\
Qwen3.5-27B       & 64.9 & 70.0 & 78.4 & 76.4 & 39.6 & 65.9 \\
InternVL3.5-4B   & 61.9 & 60.4 & 68.4 & 46.8 & 23.6 & 51.8 \\
Qwen3.5-4B        & 60.3 & 67.2 & 74.0 & 66.8 & 33.6 & 60.4 \\

\midrule

\rowcolor{sectionbg}
Random chance & 25.0 & 25.0 & 25.0 & 26.7 & 25.0 & 25.4 \\

\midrule

\rowcolor{sectionbg}
\multicolumn{7}{l}{\textit{\small Large open-source ($\geq\!27$B parameters)}} \\[1pt]
LLava-Video-32B & 60.3 & 45.0 & 42.0 & 51.0 & 28.7 & 43.8 \\
Qwen3.5-27B          & 64.9 & \textbf{\underline{67.7}} & \textbf{\underline{74.8}} & \textbf{\underline{77.3}} & \textbf{\underline{43.3}} & \textbf{\underline{68.8}} \\
Qwen3-VL-32B         & 63.4 & 64.0 & 66.9 & 68.8 & 40.5 & 62.8 \\
Gemma-4-31B           & 63.9 & 63.0 & 72.9 & 61.6 & 42.0 & 62.8 \\
Qwen2.5-VL-32B       & 66.0 & 54.4 & 70.5 & 64.7 & 34.3 & 59.3 \\
Gemma-3-27B           & 67.0 & 37.9 & 35.1 & 26.4 & 25.8 & 33.5 \\
InternVL3.5-38B   & \underline{68.0} & 65.5 & 68.4 & 68.7 & 38.6 & 63.6 \\

\midrule

\rowcolor{sectionbg}
\multicolumn{7}{l}{\textit{\small Mid-size open-source (7--9B parameters)}} \\[1pt]
Qwen3.5-9B           & 64.9 & \underline{64.6} & 69.1 & \underline{70.3} & \underline{37.6} & \underline{63.7} \\
Qwen3-VL-8B          & 61.9 & 57.4 & \underline{70.9} & 66.2 & 30.8 & 60.1 \\
InternVL3.5-8B    & 60.3 & 58.7 & 65.5 & 59.7 & 33.6 & 57.6 \\
VideoLLaMA3-7B  & \textbf{\underline{71.6}} & 51.7 & 52.6 & 58.4 & 27.3 & 50.9 \\
LLaVA-OneVision-7B    & 67.0 & 47.5 & 53.8 & 53.4 & 27.8 & 48.7 \\
LLaVA-Video-7B  & 53.6 & 34.0 & 53.9 & 39.2 & 25.0 & 40.6 \\

\midrule

\rowcolor{sectionbg}
\multicolumn{7}{l}{\textit{\small Small open-source ($\leq\!4$B parameters)}} \\[1pt]
Qwen3.5-4B            & 60.3 & \underline{66.0} & \underline{69.9} & \underline{69.0} & 35.3 & \underline{63.6} \\
Qwen3-VL-4B           & \underline{68.0} & 60.0 & 67.6 & 63.4 & 30.1 & 59.3 \\
InternVL3.5-4B    & 61.9 & 54.8 & 66.2 & 52.5 & \underline{31.6} & 54.7 \\
Gemma-4-E4B           & 52.1 & 47.9 & 64.6 & 43.2 & 28.2 & 49.2 \\
Gemma-3-4B            & 48.5 & 30.7 & 27.2 & 24.3 & 24.2 & 27.7 \\

\specialrule{1.5pt}{0pt}{0pt}  
\end{tabular}
\end{table*}


\subsection{Results}
\label{sec:results}
Table~\ref{tab:main_results} reports per-diagnostic and average
accuracy for all models and the pooled human reference.

\textbf{No model comes close to the human reference.} The pooled human reference answers 91.0\% of its 1,194 questions correctly.
On those same questions, Qwen3.5-27B scores 65.9\%,
299 correct answers behind the pooled human reference---a 25-point gap;
on the full 6,701-question benchmark, it leads at 68.8\%.
The gap is not a quirk of open-source scale: GPT-5.2, the strongest
closed-source model, scores 67.6\%, and Gemini 3 Flash reaches only 44.6\%,
barely above the applicable random baseline on several diagnostics.
\DatasetName{} is not a benchmark current VLMs are close to solving.

\textbf{Gaze detection is a wall.}
Gaze Detection (D5) is the diagnostic where models fail most uniformly and
most severely. The pooled human reference scores 89.6\%; the best model in any condition reaches
47.2\% (GPT-5.2 on the review subset), a gap of 42.4 points.  On the full
dataset, Qwen3.5-27B leads at 43.3\% and every other model falls below 40\%,
with several sitting near chance. Gaze depends on subtle, low-bandwidth
cues such as head orientation and eye-line, which the tested models do not
reliably detect. The text-only control offers no consistent shortcut
(Section~\ref{subsec:text_only}).


\textbf{Actor disambiguation (D2) is the sharpest model discriminator.}
Scores range from 34.0\% (LLaVA-Video-7B, barely above chance) to 67.7\%
(Qwen3.5-27B), a 33-point spread, larger than any other diagnostic. A paired
video-cluster bootstrap confirms a 22.7-point gap between Qwen3.5-27B and
LLaVA-Video-32B [17.7, 27.6], while its 2.25-point lead over
InternVL3.5-38B is unresolved [$-$1.51, 6.11]. The Qwen3.5
series holds together remarkably well across scales, while LLaVA-Video and Gemma-3
cluster near the chance floor.  That a 27-billion-parameter model can score
at chance on a diagnostic that a 4-billion-parameter model handles at 66\%
tells us that parameter count is not the operative variable.
A sports-video pilot also records similar trends (Appendix~\ref{app:portability}).

\begin{findingbox}{1}
Models often recognize an action but assign it to the wrong person.
For most tested models, wrong-actor answers are selected more often than
uniform distractor guessing predicts. The binding trap exposes a systematic
identity error hidden by ordinary video-question accuracy.
\end{findingbox}

\textbf{Scale helps within families but does not explain the family ranking.}
Qwen3.5-4B and InternVL3.5-38B both score 63.6\% overall despite a nearly
tenfold size difference; Qwen3.5-27B gains 5.18 points over its 4B version,
while Gemma-4-E4B exceeds Gemma-3-27B by 15.73. On actor disambiguation,
Qwen3.5 changes little across 4--27B, while InternVL3.5 and LLaVA-Video
each gain about 11 points with scale. Qwen3.5 and InternVL3.5 document
spatial-grounding evaluations
\cite{qwen2026qwen35card,wang2025internvl35};
LLaVA-Video's disclosed video data emphasize captioning and QA, and its
adapter pools spatial tokens \cite{zhang2024llavavideo}. Training and input
differences co-vary with model generation, so this comparison does not
identify a single cause (Appendix~\ref{app:model_family}).

\begin{findingbox}{2}
Parameter count alone does not predict actor-binding performance. A 4B
model scores alongside a 38B model overall, while larger variants deliver clear gains
within some families. Size is an unreliable shortcut for comparing them.
\end{findingbox}

\noindent\textbf{The large gaps survive video-level resampling.}
In 10,000 paired bootstrap resamples of the 64 source videos, Qwen3.5-27B
leads InternVL3.5-38B by 5.18 points in question-micro accuracy
(95\% CI [3.61, 6.75]) and LLaVA-Video-32B by 24.98 points
([22.61, 27.35]). Qwen3.5-4B and InternVL3.5-38B, by contrast,
are tied to reported precision (difference 0.00 points;
[-1.74, 1.59]). The family-scale and cross-family gaps are real;
small leaderboard separations need not be. Full intervals and video-level
robustness checks appear in Appendix~\ref{app:statistical_uncertainty}.


\begin{findingbox}{3}
Aggregate video accuracy hides uneven abilities. Models can detect
transitions and concurrent actions while struggling to bind actions to the
right person or read gaze. The pooled human reference remains far ahead on
these fine-grained judgments.
\end{findingbox}

COT prompting does not improve accuracy and can
sharply hurt interaction reasoning (Appendix~\ref{app:prompting}).

\begin{figure*}[t]
  \centering
  \begin{subfigure}[b]{0.63\linewidth}
    \centering
    \includegraphics[width=\linewidth]{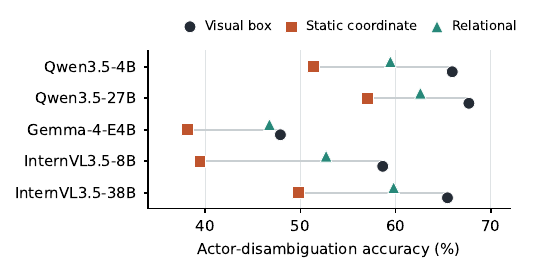}
    \caption{%
      \textbf{Who is the target?} On actor disambiguation, relational references
      recover much of the static-coordinate penalty for all five models;
      tracked visual boxes remain strongest.
    }
    \label{fig:reference_interfaces}
  \end{subfigure}
  \hfill
  \begin{subfigure}[b]{0.35\linewidth}
    \centering
    \includegraphics[width=\linewidth]{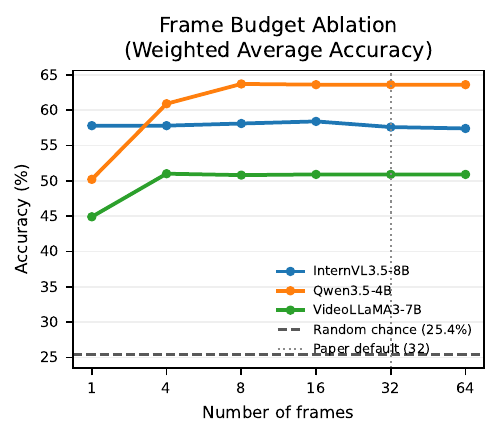}
    \caption{%
      \textbf{Frame budget.} Accuracy vs.\ frame count.
      Gains saturate by 8--16 frames; the 32-frame default adds little.
    }
    \label{fig:frame_sweep}
  \end{subfigure}
  \caption{%
    \textbf{Reference and temporal controls.} The actor-reference interface
    changes accuracy (left); additional frames offer little gain beyond 8--16 (right).%
  }
  \label{fig:ablations}
\end{figure*}

\subsection{Reference Interface Is an Actor-Binding Bottleneck}
\label{subsec:grounding}

We evaluate five models — InternVL3.5-8B, InternVL3.5-38B, Qwen3.5-4B,
Qwen3.5-27B, and Gemma-4-E4B — on all 2,000 actor-disambiguation and 1,611
interaction-reasoning questions with three reference interfaces. The \emph{visual-box}
condition uses tracked overlays. The \emph{static-coordinate reference}
condition uses the raw unboxed video and the target box at the annotated
midpoint, with no frame or timestamp token; coordinates are normalised to
$[0,1000]$ for Qwen and InternVL and $[0,1]$ for other models. The
\emph{relational reference} condition also uses raw unboxed video but names
the target with a deterministic ordinary-language description at the midpoint,
such as ``leftmost person'' or ``second person from the left.''

\noindent\textbf{Numeric coordinates impose a large actor-disambiguation penalty.}
Figure~\ref{fig:reference_interfaces} shows the three interfaces on the same
questions. Actor-disambiguation accuracy falls by 9.8--19.2 points when tracked
boxes are replaced with static coordinates. For InternVL3.5-8B, it falls from
58.65\% to 39.45\%. But this
condition is not information-equivalent to a tracked overlay: it requires the
model to parse a numeric convention and associate one midpoint box with a
person throughout the clip. The coordinate-only contrast therefore does not
identify a pure spatial-grounding deficit, but it does establish that numeric
coordinates are a poor actor-reference interface for all five models.

\noindent\textbf{Relational language recovers a substantial share.}
Across all five models, relational descriptions improve over static coordinates
by 5.55--13.25 points on actor disambiguation, while remaining 1.15--6.50 points
below visual boxes for every model. InternVL3.5-8B, for example, recovers to
52.70\% with a relational reference. Numeric-coordinate parsing therefore explains a material
share of the original drop, but it does not explain the full advantage of an
explicit tracked visual anchor.

\noindent\textbf{Interaction reasoning exposes a harder two-reference regime.}
Relational descriptions improve over static coordinates for Gemma-4-E4B
(+10.09 points) and InternVL3.5-8B (+9.11), but are similar or worse for the
other three models. Visual boxes lead for four models; Qwen3.5-27B instead
scores highest with coordinates (79.58\%). Resolving two people produces no
universal ordering of unboxed references. The exact results for both
diagnostics appear in Appendix~\ref{app:reference_interfaces}.

\begin{findingbox}{4}
How the target is specified changes what models can answer. Relational
language recovers much of the loss from static coordinates on actor
disambiguation, yet tracked visual boxes still lead for every tested model.
Two-person interaction reasoning has no universally best unboxed format.
\end{findingbox}

Swapping the target-box color yields no consistent red or blue advantage
across models (Appendix~\ref{app:color_swap}).

\subsection{What's the optimal Frame Budget}
\label{subsec:frame_budget}

We sweep frame counts $\{1, 4, 8, 16, 32, 64\}$ on InternVL3.5-8B,
Qwen3.5-4B, and VideoLLaMA3-7B to understand how much temporal information
each diagnostic actually requires.  Figure~\ref{fig:frame_sweep} shows the
weighted-average accuracy; per-diagnostic curves are in
Appendix~\ref{app:frame_sweep}.

\noindent\textbf{Rapid saturation on the weighted average.}
All three models improve substantially from 1 to 8 frames
then plateau.  Doubling the budget from 32 to 64 frames yields essentially
no gain for any model, confirming that 32 frames is a sufficient operating
point for the main evaluation.

\noindent\textbf{Transition sensitivity is the most frame-hungry diagnostic.}
Transition Sensitivity requires detecting that an action \emph{changed},
and the 1-frame baseline reflects this directly: InternVL3.5-8B drops to
40.2\% with a single frame, recovering to 63.4\% by
8 frames. VideoLLaMA3-7B also shows a sharp transition-sensitivity recovery from 58.3\% (1
frame) to 71.7\% (4 frames).  Temporal sequence is not optional for this
diagnostic.

\noindent\textbf{Concurrent binding is the least frame-hungry.}
Concurrent Action Binding peaks early: InternVL3.5-8B achieves 71.4\% with
a single frame and does not improve
meaningfully with more.  Recognising that a person is simultaneously
performing two actions is largely solvable from a single well-chosen frame,
consistent with the ``snapshot'' nature of co-occurrence annotations.

\noindent\textbf{InternVL3.5-8B is frame-agnostic; Qwen3.5-4B is
frame-sensitive.}
InternVL3.5-8B's weighted average barely moves across all budgets (57.8\% at 1 frame, 57.6\% at 32).
In contrast, Qwen3.5-4B gains 13.5 points from 1 to 8 frames, driven by actor disambiguation ($+$25.3) and interaction reasoning ($+$10.7) ---
precisely the person-binding diagnostics that require seeing more of the scene.

\subsection{Can Models Solve ActionLens from Text Alone?}
\label{subsec:text_only}

\noindent\textbf{Video carries most of the signal.}
Removing the clip lowers question-micro accuracy by 15.2--33.4 points across
five models (Appendix~\ref{app:text_only}). Actor disambiguation, concurrent
binding, and interaction reasoning decline for every model: their answers
cannot be recovered reliably from the question text. Transition sensitivity
is less clean. Familiar action sequences can help models guess a
``before/after'' answer, and one model even improves slightly without video.
Gaze remains difficult with video available; seeing the clip alone does not
resolve this subtle cue.

\begin{findingbox}{5}
Seeing the clip matters; adding frames alone has limited returns. Removing
video lowers accuracy for every tested model, while the frame-sweep models
gain little beyond 8--16 frames. More temporal sampling does not erase the
binding failures.
\end{findingbox}

\section{Conclusion}
\DatasetName{} asks a simple question: can VLMs watch two people and say what each is doing?
After 6,701 questions, 20 models, and 14 rounds of quality engineering, the answer is: not really.
On the same 1,194 questions, the best open-weight model sits 25 points below the pooled human reference; gaze detection hovers near chance; and 13 of 16 models are more likely to select the \emph{wrong} actor's action than to guess randomly.
The reference control sharpens the diagnosis: numeric parsing causes a substantial share of the actor-disambiguation drop, yet every model still trails tracked visual boxes with ordinary relational descriptions. For interaction reasoning, no single unboxed reference format wins across models.
These are not quirks of dataset construction.
They are named, measurable sub-problems: person tracking, multi-actor binding, and gaze inference.
\DatasetName{} makes each one easy to quantify. The dataset, code, and a one-command script to run any \texttt{lmms-eval}-supported model are publicly available at \url{https://anonymous.4open.science/r/lmms-eval-2276}.

\subsection*{AI use statement}
Language models assisted with text editing, manuscript review, and code for auditing experimental outputs. The authors verified AI-assisted code and all reported numerical results against the underlying data and take responsibility for the scientific claims, interpretations, and final content.

\subsection*{Ethics statement}
ActionLens derives questions from AVA~v2.2's movie footage and person-action
annotations~\cite{gu2018ava}. AVA lists its dataset under a CC BY 4.0 license
(\url{https://sites.research.google/gr/ava/download/}); use of the underlying
footage should follow its applicable source terms. Four non-author raters
provided the pooled human reference, reported as aggregate accuracy.
ActionLens is intended to diagnose models, not to assess the people depicted.
Movie casting and editing limit whose actions and contexts are represented;
benchmark scores should not be used to infer suitability for surveillance or
other consequential decisions.

\subsection*{Reproducibility statement}
We release the ActionLens dataset, construction and analysis code, task configurations, prompts, raw evaluation outputs, and scripts needed to reproduce the reported results at \url{https://anonymous.4open.science/r/lmms-eval-2276}. Appendix~\ref{app:repro} documents model settings, video sampling, metrics, human-reference collection and clustered uncertainty analyses.

\bibliography{iclr2027_conference}
\bibliographystyle{iclr2027_conference}

\clearpage
\appendix
\section{Appendix}
\section{Full Experimental Setup}
\label{app:setup}

\subsection{Models}

Table~\ref{tab:app_models} lists all evaluated models with their parameter
counts and evaluation scope.

\begin{table}[h]
\centering\small
\setlength{\tabcolsep}{4pt}
\renewcommand{\arraystretch}{1.12}
\caption{\textbf{All evaluated models.} ``Full'' = full 6,701-item dataset;
``Subset'' = 1,194-item human-review subset.}
\label{tab:app_models}
\begin{tabular}{l r l}
\toprule
\textbf{Model} & \textbf{Params} & \textbf{Scope} \\
\midrule
\multicolumn{3}{l}{\textit{Closed-source}} \\
GPT-5.2~\cite{openai2026gpt5}             & -- & Subset \\
Gemini 3 Flash~\cite{team2026gemini3flash} & -- & Subset \\
\midrule
\multicolumn{3}{l}{\textit{Large open-source ($\geq$27B)}} \\
Qwen3.5-27B~\cite{qwen2026qwen35}          & 27B & Full \\
InternVL3.5-38B~\cite{wang2025internvl35}  & 38B & Full \\
Qwen3-VL-32B~\cite{bai2025qwen3vl}         & 32B & Full \\
Qwen2.5-VL-32B~\cite{bai2025qwen25vl}      & 32B & Full \\
Gemma-4-31B~\cite{team2026gemma4}           & 31B & Full \\
Gemma-3-27B~\cite{team2025gemma3}           & 27B & Full \\
LLaVA-Video-32B~\cite{zhang2024llavavideo}  & 32B & Full \\
\midrule
\multicolumn{3}{l}{\textit{Mid-size open-source (7--9B)}} \\
Qwen3.5-9B~\cite{qwen2026qwen35}            & 9B  & Full \\
Qwen3-VL-8B~\cite{bai2025qwen3vl}           & 8B  & Full \\
InternVL3.5-8B~\cite{wang2025internvl35}    & 8B  & Full \\
VideoLLaMA3-7B~\cite{zhang2025videollama3}  & 7B  & Full \\
LLaVA-OneVision-7B~\cite{li2024llavaov}    & 7B  & Full \\
LLaVA-Video-7B~\cite{zhang2024llavavideo}   & 7B  & Full \\
\midrule
\multicolumn{3}{l}{\textit{Small open-source ($\leq$4B)}} \\
Qwen3.5-4B~\cite{qwen2026qwen35}            & 4B  & Full \\
Qwen3-VL-4B~\cite{bai2025qwen3vl}           & 4B  & Full \\
InternVL3.5-4B~\cite{wang2025internvl35}    & 4B  & Full \\
Gemma-4-E4B~\cite{team2026gemma4}            & 4B  & Full \\
Gemma-3-4B~\cite{team2025gemma3}             & 4B  & Full \\
\bottomrule
\end{tabular}
\end{table}

\subsection{Prompt Template}

All models share one template — no chain-of-thought, no few-shot examples,
no task-specific system prompt:

\begin{quote}\small\tt
\{question\}\\
A.\ \{option\_A\} \quad B.\ \{option\_B\} \quad C.\ \{option\_C\} \quad D.\ \{option\_D\}\\
Answer with the option letter only.
\end{quote}

\noindent Model-specific chat templates are applied internally by
\texttt{lmms-eval}. For the 21\% of Interaction Reasoning items with only three options,
option~D is omitted from the prompt.

\subsection{Video Sampling and Hardware}

Clips are sampled uniformly to \textbf{32 frames} by default.
Models up to 32B run on a single \textbf{NVIDIA H100 80\,GB SXM5} in
\texttt{bfloat16}.  InternVL3.5-38B uses two H100s with tensor parallelism.
Full-dataset runs take 1--2.5 hours; the full experiment matrix is
$\approx$50 GPU-hours.

\subsection{Metric Details}

A response is correct iff the first predicted option letter matches the
ground-truth label. The primary aggregate, \emph{question-micro accuracy}, is
the fraction correct across all 6,701 full-set questions:
$\sum_d N_d \cdot \text{acc}_d \;/\; \sum_d N_d$,
with $N_{\text{D1}}\!=\!194$, $N_{\text{D2}}\!=\!N_{\text{D3}}\!=\!2{,}000$,
$N_{\text{D4}}\!=\!1{,}611$, $N_{\text{D5}}\!=\!896$.
The pooled human reference and closed-source models use the same definition
on their 1,194-item review subset. We also report the mean of the five
diagnostic accuracies separately as \emph{diagnostic-macro accuracy} in
Appendix~\ref{app:statistical_uncertainty}; this gives each diagnostic equal weight.

\section{Quality Engineering: The 14 Rounds}
\label{app:quality}

We started from a 281-sample pilot with 53\% human accuracy and ended above
90\% through 14 targeted review-and-fix iterations.
Table~\ref{tab:fix_rounds} summarises every round.
Two diagnostics were retired during this process (Section~\ref{app:retired}).

\begin{table*}[h]
\centering\small
\setlength{\tabcolsep}{5pt}
\renewcommand{\arraystretch}{1.18}
\caption{\textbf{14-round quality-engineering progression.}
``Acc'' = human accuracy on the reviewed sample after fixes from the
\emph{previous} round. Fix numbers correspond to the detailed log below.}
\label{tab:fix_rounds}
\begin{tabular}{c r r l l}
\toprule
\textbf{Round} & \textbf{Samples} & \textbf{Acc (\%)} &
\textbf{Primary failure mode} & \textbf{Fix(es)} \\
\midrule
Pre  & 281 & 53 & Multiple simultaneous failures & --- \\
1--2 & 54  & 74 & Distractor contamination; auditory actions; degenerate gaze labels; filler distractors & 1--6 \\
3    & 50  & 74 & Track drift; confusable pairs; shared co-actions; overlapping boxes & 7--12 \\
4    & 100 & 79 & Boundary detection; crowd confounders & 13 \\
5    & 100 & 82 & Residual auditory leakage in D1 & 14 \\
6    & 100 & 84 & Segment centring for D1 & 15 \\
7    & 150 & 86 & D4 unique co-action requirement & 16 \\
8    & 150 & 87 & GT person-ID tracking deployed & 17 \\
9    & 200 & 88 & Max bbox area filter & 18 \\
10   & 200 & 89 & IoU pair-overlap guard & 19 \\
11   & 250 & 90 & Answer-balance audit; D4 3-option subset & 20 \\
12   & 250 & 91 & Locally-stable boundary scan (D1) & 21 \\
13   & 300 & 91 & Physically-compatible distractor pairs (D3) & 22 \\
14   & 1,194 & \textbf{91.0} & Pooled human reference — no further fixes & --- \\
\bottomrule
\end{tabular}
\end{table*}

\subsection{Detailed Fix Log}

\paragraph{Fix 1 — Gaze Detection (D5): Degenerate gaze-type distribution.}
100\% of initial candidates had \texttt{gaze\_type = red\_watches\_blue}.
Fix: emit all four categories (\emph{red watches blue}, \emph{blue watches
red}, \emph{mutual}, \emph{neither}) with balanced sampling ($\approx$25\%
each).

\paragraph{Fix 2 — Actor Disambiguation (D2): Distractor contamination.}
85.5\% of questions had at least one distractor that was a true action of
the target person.  Fix: the distractor sampler receives the full set of true
actions as an \texttt{excluded} set.

\paragraph{Fix 3 — Concurrent Binding (D3): Multi-correct action pairs.}
57.1\% of candidates had 3+ simultaneous actions, making several distractor
pairs also valid.  Fix: restrict to time-steps with \emph{exactly} two visual
actions.

\paragraph{Fix 4 — Transition Sensitivity (D1): Auditory-only transitions.}
61.1\% of candidates were \texttt{talk\,$\leftrightarrow$\,listen}, invisible
in muted video.  Fix: skip transitions where both actions are in
\texttt{AUDITORY\_ONLY\_ACTIONS}.

\paragraph{Fix 5 — Actor Disambiguation (D2): Auditory correct answers.}
54\% of correct answers were auditory actions. Fix: require non-empty
visual-only distinct action set per person.

\paragraph{Fix 6 — Interaction Reasoning (D4): Generic filler distractors.}
Distractors used fixed phrases trivially distinguishable from factual correct
answers.  Fix: all four options are structured variants (correct, role-swap,
co-action-swap, scene-alternative).  When the scene pool is too small to
furnish a valid scene-alternative, the question is emitted as a 3-option item
(21\% of Interaction Reasoning items); option~D is omitted from the prompt in that case.

\paragraph{Fix 7 — Transition Sensitivity (D1): Track identity drift.}
IoU tracker (threshold 0.7) drifted across nearby people; 54.5\% of items
were marked unanswerable.  Fix: reject tracks whose bbox centre jumps
$>$15\% of the frame between consecutive seconds.

\paragraph{Fix 8 — Transition Sensitivity (D1): Visually confusable pairs.}
\texttt{ride\,$\leftrightarrow$\,drive} transitions are visually
indistinguishable.  Fix: \texttt{CONFUSABLE\_TRANSITION\_PAIRS} blacklist.

\paragraph{Fix 9 — Interaction Reasoning (D4): Shared co-actions produce multi-correct options.}
When both persons share the same co-actions, role-swap distractors remain
valid.  Fix: both persons must have \emph{unique} visual co-actions not shared
by the other.

\paragraph{Fix 10 — Gaze Detection (D5): Distant pairs and third-person confounders.}
AVA's \texttt{watch (a person)} does not specify who is watched.  In crowded
scenes the target may be unboxed.  Fix: proximity filter ($\leq$0.35
normalised) + maximum 2 people in scene.

\paragraph{Fix 11 — All: Maximum bounding-box area filter.}
Oversized detections ($>$60\% of frame) cause tracker drift.  Fix: drop such
instances before track building.

\paragraph{Fix 12 — All: Ground-truth person-ID tracking.}
AVA's CSV column~8 is \texttt{person\_id}, not \texttt{label\_confidence}.
The pipeline was building its own IoU tracker as a result.  Fix: group
observations directly by \texttt{person\_id}; eliminates identity drift at
the source and supersedes Fixes 7 and 11.

\paragraph{Fix 13 — Transition Sensitivity (D1): Locally-stable boundary scan.}
Long tracks produced noisy endpoint-to-endpoint transition labels.  Fix: a
boundary qualifies only when the old action is stable ($\geq$2 consecutive
seconds) before, and the new action stable after; segment is centred on the
boundary midpoint and extended to the stable-region limits, yielding
variable-length clips (4--10\,s, mean 7.1\,s).

\paragraph{Fixes 14--22 — Rounds 5--13.}
Subsequent rounds addressed residual auditory leakage, Transition Sensitivity segment-centring
edge cases, the Interaction Reasoning unique co-action requirement at scale, answer-balance audits
(maximum 5.1\,pp deviation from the 25\% four-position reference), and physically-compatible
distractor pairs for Concurrent Binding via \texttt{MUTUALLY\_EXCLUSIVE\_GROUPS}.

\subsection{Retired Diagnostics}
\label{app:retired}

\paragraph{Talker/Listener Detection.}
Pre-revision human accuracy was 97.3\% on a binary question grounded in AVA
talk/listen labels. As a role-only test, it could depend on speech cues absent
from muted video and was retired. Interaction Reasoning (D4) uses the same
labels only in composite choices with distinct visible co-actions; its result
should not be read as isolated speaker-identification accuracy. The composite
format does not independently validate every communication-role label from
silent video.

\paragraph{Temporal Localisation.}
Localising an action onset is fundamentally ambiguous for MCQ without a
tolerance window; any nearby second is arguably correct.

\section{Dataset Composition}
\label{app:dataset_stats}

\begin{table}[h]
\centering\small
\setlength{\tabcolsep}{5pt}
\renewcommand{\arraystretch}{1.15}
\caption{\textbf{Per-diagnostic statistics.}
$^\dagger$D1 clips have variable duration (4--10\,s, mean 7.1\,s).
$^\ddagger$D4 contains 334 three-option items (21\%).}
\label{tab:app_dataset_stats}
\begin{tabular}{c l r r c r c}
\toprule
\textbf{ID} & \textbf{Diagnostic} & \textbf{N} & \textbf{Videos} & \textbf{Duration} & \textbf{People} & \textbf{Options} \\
\midrule
D1 & Transition Sensitivity   & 194   & 53 & 4--10\,s$^\dagger$ & 1    & 4 \\
D2 & Actor Disambiguation     & 2,000 & 63 & 5\,s               & 2--7 & 4 \\
D3 & Concurrent Binding       & 2,000 & 64 & 5\,s               & 1    & 4 \\
D4 & Interaction Reasoning$^\ddagger$ & 1,611 & 61 & 5\,s      & 2--7 & 3--4 \\
D5 & Gaze Detection           & 896   & 62 & 5\,s               & 2    & 4 \\
\midrule
   & \textbf{Total}           & \textbf{6,701} & \textbf{64} & & & \\
\bottomrule
\end{tabular}
\end{table}

\begin{figure}[h]
  \centering
  \includegraphics[width=0.7\linewidth]{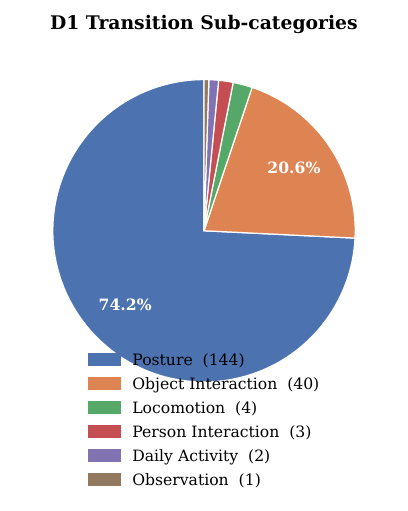}
  \caption{\textbf{Transition Sensitivity type distribution.}
    Posture transitions (e.g., stand\,$\to$\,bend/bow) are the most common;
    object-interaction and locomotion transitions are also well represented.
    ``Other'' covers single-item categories below 1\%.}
  \label{fig:app_d1_transitions}
\end{figure}

\paragraph{Transition Sensitivity (D1) type breakdown.}
Figure~\ref{fig:app_d1_transitions} shows the distribution of transition
categories. Posture changes dominate (74\%), followed by
object-interaction changes (21\%), locomotion changes (2\%), and others (3\%).
This distribution reflects AVA's annotation density rather than a design
choice; it confirms that this diagnostic is not dominated by a single trivially recognisable
pattern.

\paragraph{Interaction Reasoning (D4) distractor design.}
Each question asks how two highlighted people are interacting (e.g., ``Person A talks while throwing; Person B listens while watching'').
Four structured distractor types are generated:
\emph{role swap} (talker and listener roles exchanged),
\emph{co-action swap} (unique co-actions swapped between persons),
\emph{scene alternative} (same talker/listener structure but with a different action drawn from the scene pool), and
a \emph{3-option variant} (21\% of items) that omits the scene alternative when the scene pool contains fewer than two distinct alternatives.
The unique co-action requirement (Fix~9) is essential: if both persons shared the same co-actions, the role-swap distractor would remain factually correct, making the question multi-answer.

\paragraph{Gaze Detection (D5) construction.}
AVA's \texttt{watch (a person)} label is a per-person binary annotation: it records \emph{whether} a person is watching someone, not \emph{whom}.
Rather than claiming directional gaze (which would require knowing the watching target), this diagnostic asks the weaker but cleanly answerable question of \emph{detection per person}.
A \emph{3-second temporal consistency} filter requires the gaze label to agree for at least 3 of the 5 clip seconds, ensuring the model can answer from the majority of visible frames.
After filtering (5,353 raw candidates $\to$ 1,422 consistent), balanced sampling yields 224 items per gaze type.

\begin{figure}[h]
  \centering
  \includegraphics[width=0.6\linewidth]{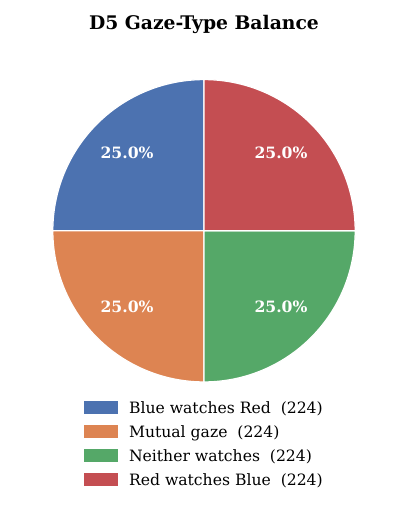}
  \caption{\textbf{Gaze Detection category balance.}
    All four categories (red watches blue, blue watches red, mutual, neither)
    fall within 1.5\,pp of the 25\% uniform baseline.}
  \label{fig:app_d5_gaze}
\end{figure}

\paragraph{Gaze Detection category balance.}
Figure~\ref{fig:app_d5_gaze} confirms the near-perfect balance of the four
gaze categories after Fix~1 and balanced sampling.  No single answer option
can be exploited as a positional shortcut.

\section{Static-Coordinate Reference: Full Results}
\label{app:coord_full}

\begin{figure*}[h]
  \centering
  \includegraphics[width=\linewidth]{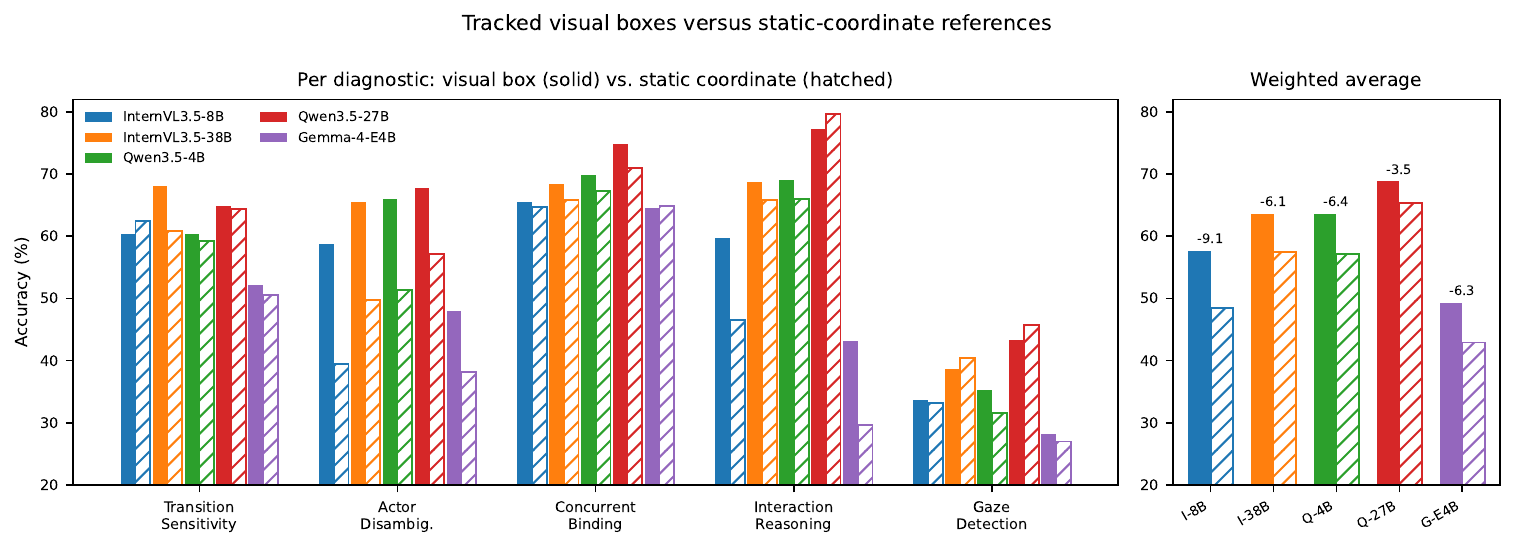}
  \caption{\textbf{Visual-box versus static-coordinate reference — full view.}
    Solid bars: tracked visual boxes. Hatched bars: static midpoint coordinates.
    \emph{Left}: per-diagnostic breakdown; Actor Disambiguation decreases for every model.
    \emph{Right}: weighted-average summary with delta annotations.
    This two-interface comparison includes coordinate-format sensitivity and
    should not be interpreted as a pure grounding measure.}
  \label{fig:app_coord_combined}
\end{figure*}

Figure~\ref{fig:app_coord_combined} shows the full static-coordinate reference comparison
across all five models and all diagnostics — both the per-diagnostic accuracy
bars and the weighted-average summary.

Table~\ref{tab:app_coord} gives the exact numbers.

\begin{table}[h]
\centering\small
\setlength{\tabcolsep}{3.5pt}
\renewcommand{\arraystretch}{1.15}
\caption{\textbf{Visual-box versus static-coordinate reference — full numbers.}
Each cell: visual box / static coordinate (delta in pp). \textbf{Bold}: $|\Delta|>5$.}
\label{tab:app_coord}
\begin{tabular}{l r r r r r r}
\toprule
\textbf{Model} & \textbf{D1} & \textbf{D2} & \textbf{D3} & \textbf{D4} & \textbf{D5} & \textbf{WAvg} \\
\midrule
InternVL3.5-8B
  & 60.3/62.4 \small(+2.1)
  & \textbf{58.7/39.5 \small($-$19.2)}
  & 65.5/64.7 \small($-$0.8)
  & \textbf{59.7/46.5 \small($-$13.2)}
  & 33.6/33.2 \small($-$0.4)
  & \textbf{57.6/48.5 \small($-$9.1)} \\
InternVL3.5-38B
  & \textbf{68.0/60.8 \small($-$7.2)}
  & \textbf{65.5/49.8 \small($-$15.7)}
  & 68.4/65.8 \small($-$2.6)
  & 68.7/65.8 \small($-$2.9)
  & 38.6/40.4 \small(+1.8)
  & \textbf{63.6/57.5 \small($-$6.1)} \\
Qwen3.5-4B
  & 60.3/59.3 \small($-$1.0)
  & \textbf{66.0/51.4 \small($-$14.6)}
  & 69.9/67.3 \small($-$2.6)
  & 69.0/66.0 \small($-$3.0)
  & 35.3/31.6 \small($-$3.7)
  & \textbf{63.6/57.2 \small($-$6.4)} \\
Qwen3.5-27B
  & 64.9/64.4 \small($-$0.5)
  & \textbf{67.7/57.1 \small($-$10.6)}
  & 74.8/70.9 \small($-$3.9)
  & 77.3/79.6 \small(+2.3)
  & 43.3/45.7 \small(+2.4)
  & \textbf{68.8/65.3 \small($-$3.5)} \\
Gemma-4-E4B
  & 52.1/50.5 \small($-$1.6)
  & \textbf{47.9/38.1 \small($-$9.8)}
  & 64.6/64.8 \small(+0.2)
  & \textbf{43.2/29.7 \small($-$13.5)}
  & 28.2/27.0 \small($-$1.2)
  & \textbf{49.2/42.9 \small($-$6.3)} \\
\bottomrule
\end{tabular}
\end{table}

\section{Relational Reference Control}
\label{app:reference_interfaces}

Table~\ref{tab:app_reference_interfaces} gives the paired three-interface results
for the same five models and question sets shown in Section~\ref{subsec:grounding}.
The static-coordinate and relational conditions both use raw, unboxed video;
only the target description changes. Relational references identify people by
their left-to-right position at the annotated midpoint, such as ``leftmost
person'' or ``second person from the left.''

\begin{table}[h]
\centering\small
\setlength{\tabcolsep}{3.3pt}
\caption{\textbf{Reference-interface accuracy (\%).} Each model answers all
2,000 Actor Disambiguation questions and all 1,611 Interaction Reasoning
questions under each interface. Box = tracked visual overlay; Coord = static
midpoint coordinates; Rel = midpoint relational description on unboxed video.}
\label{tab:app_reference_interfaces}
\begin{tabular}{l r r r r r r}
\toprule
& \multicolumn{3}{c}{\textbf{Actor Disambiguation}} & \multicolumn{3}{c}{\textbf{Interaction Reasoning}} \\
\cmidrule(lr){2-4}\cmidrule(l){5-7}
\textbf{Model} & \textbf{Box} & \textbf{Coord} & \textbf{Rel} & \textbf{Box} & \textbf{Coord} & \textbf{Rel} \\
\midrule
Qwen3.5-4B       & 65.95 & 51.35 & 59.45 & 69.03 & 66.03 & 65.43 \\
Qwen3.5-27B      & 67.70 & 57.05 & 62.60 & 77.28 & 79.58 & 73.00 \\
Gemma-4-E4B      & 47.90 & 38.10 & 46.75 & 43.20 & 29.70 & 39.79 \\
InternVL3.5-8B   & 58.65 & 39.45 & 52.70 & 59.65 & 46.45 & 55.56 \\
InternVL3.5-38B  & 65.45 & 49.75 & 59.80 & 68.65 & 65.80 & 64.49 \\
\bottomrule
\end{tabular}
\end{table}

\paragraph{Crowd size.}
The interaction task requires resolving two references. In the reported
crowd-size stratification, relational references trail visual boxes by
0.8--4.1 points in two-person scenes and by 8.1--13.1 points in four-person
scenes across the evaluated models. This is consistent with a growing cost of
resolving two people as scenes become more crowded.

\section{Model Families and Input Handling}
\label{app:model_family}

Table~\ref{tab:model_family_metadata} places the family differences in the
context of public model documentation. The grounding column distinguishes
reported evaluations or training data from fully specified supervision;
neither a RefCOCO score nor an unlisted task proves which examples entered
pre-training. The coordinate ranges describe \emph{our} static-coordinate
control, not a model's required native input format. The main full-set family
comparisons use the same 32-frame cap, but visual encoding differs by family.

\begin{table}[h]
\centering\footnotesize
\setlength{\tabcolsep}{3pt}
\caption{\textbf{Documented family differences relevant to actor binding.}
Grounding supervision is marked unspecified where the public sources do not
identify it. Dates denote model-family releases; the coordinate column
records our ablation convention.}
\label{tab:model_family_metadata}
\begin{tabular}{p{0.95in}p{1.50in}p{1.66in}c}
\toprule
\textbf{Family (year)} & \textbf{Grounding evidence disclosed} &
\textbf{Video/frame handling} & \textbf{Coord.} \\
\midrule
Qwen3.5 (2026)\newline 4/9/27B
  & Spatial/RefCOCO evaluations; exact grounding-training mix unspecified
  & Native video processor
  & $[0,1000]$ \\
InternVL3.5 (2025)\newline 4/8/38B
  & RefCOCO evaluations and large multimodal SFT; exact grounding-training mix unspecified
  & 448-pixel frame tiles in our evaluation
  & $[0,1000]$ \\
LLaVA-Video (2024)\newline 7/32B
  & Disclosed 178K video set contains captions and QA; box labels not listed
  & Spatial average pooling, stride 2 in our evaluation
  & $[0,1]$ \\
Gemma 3 (2025)\newline 27B
  & Grounding-task supervision unspecified
  & 896-pixel image inputs, 256 tokens per image
  & $[0,1]$ \\
Gemma 4 (2026)\newline E4B
  & Grounding-task supervision unspecified; object detection and pointing described
  & Video as image frames; configurable visual token budget
  & $[0,1]$ \\
\bottomrule
\end{tabular}
\end{table}

The sources are the Qwen3.5 release and model card
\cite{qwen2026qwen35,qwen2026qwen35card}, the InternVL3.5 report
\cite{wang2025internvl35}, the LLaVA-Video report
\cite{zhang2024llavavideo}, and the Gemma 3 and 4 model cards
\cite{gemmateam2025gemma3card,team2026gemma4}. Our released evaluation
adapters specify the InternVL tile size and LLaVA-Video pooling setting.
Public training descriptions
do not establish whether AVA or ActionLens-like tasks were included.
These architectural and training differences co-vary with model generation,
so the table motivates hypotheses rather than attributing the observed
family gaps to any single cause.

\paragraph{Actor-disambiguation scaling.}
On the same 2,000 questions, Qwen3.5-4B/9B/27B score
65.95/64.55/67.70\%, respectively; the 27B--4B paired video-cluster
interval [$-$1.34, 4.66] spans zero. InternVL3.5-4B/8B/38B score
54.75/58.65/65.45\%, with a 10.70-point 38B--4B gap
[6.67, 14.71]. LLaVA-Video-7B/32B rises from 34.05\% to 45.00\%,
a 10.95-point gap [6.10, 16.08]. Across Gemma generations,
Gemma-4-E4B exceeds Gemma-3-27B by 10.05 points on actor
disambiguation [4.25, 15.71] and 15.73 points in full-set
question-micro accuracy [12.67, 18.81]. All intervals use 10,000
paired bootstrap resamples of source videos; the actor-disambiguation
analysis clusters its 2,000 questions by 63 source videos.

\section{Binding Trap Analysis: Full Results}
\label{app:binding_trap}

\begin{figure}[htbp]
  \centering
  \includegraphics[width=0.82\linewidth]{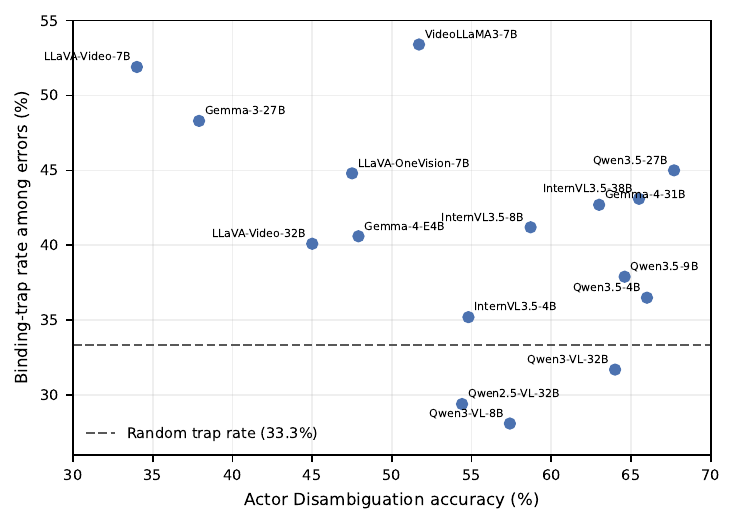}
  \caption{\textbf{Actor Disambiguation accuracy vs.\ binding-trap rate.}
    Each point is one model. The trap rate is the fraction of incorrect
    responses that selected the binding-trap distractor (the other person's
    correct action).  Random expectation: 33.3\%.  Models above the dashed
    line are systematically biased toward the wrong actor's action.}
  \label{fig:app_binding_scatter}
\end{figure}

Figure~\ref{fig:app_binding_scatter} plots Actor Disambiguation accuracy against trap rate for
all 16 models, complementing the bar chart in the main paper.
The negative correlation is clear: models that fall into the trap most often
are also the weakest on Actor Disambiguation overall.

Table~\ref{tab:app_binding_trap} gives the full numerical breakdown.

\begin{table}[h]
\centering\small
\setlength{\tabcolsep}{5pt}
\renewcommand{\arraystretch}{1.15}
\caption{\textbf{Binding-trap rates for Actor Disambiguation.}
Trap rate = fraction of incorrect responses landing on the binding-trap
option. Random expectation: 33.3\%. Lift $>$1.0 = systematic actor-binding
failure.}
\label{tab:app_binding_trap}
\begin{tabular}{l r r r}
\toprule
\textbf{Model} & \textbf{Actor Acc. (\%)} & \textbf{Trap rate (\%)} & \textbf{Lift} \\
\midrule
VideoLLaMA3-7B     & 51.7 & 53.4 & 1.60$\times$ \\
LLaVA-Video-7B     & 34.0 & 51.9 & 1.56$\times$ \\
Gemma-3-27B        & 37.9 & 48.3 & 1.45$\times$ \\
Qwen3.5-27B        & 67.7 & 45.0 & 1.35$\times$ \\
LLaVA-OneVision-7B & 47.5 & 44.8 & 1.34$\times$ \\
InternVL3.5-38B    & 65.5 & 43.1 & 1.29$\times$ \\
Gemma-4-31B        & 63.0 & 42.7 & 1.28$\times$ \\
InternVL3.5-8B     & 58.7 & 41.2 & 1.24$\times$ \\
Gemma-4-E4B        & 47.9 & 40.6 & 1.22$\times$ \\
LLaVA-Video-32B    & 45.0 & 40.1 & 1.20$\times$ \\
Qwen3.5-9B         & 64.6 & 37.9 & 1.14$\times$ \\
Qwen3.5-4B         & 66.0 & 36.5 & 1.10$\times$ \\
InternVL3.5-4B     & 54.8 & 35.2 & 1.06$\times$ \\
Qwen3-VL-32B       & 64.0 & 31.7 & 0.95$\times$ \\
Qwen2.5-VL-32B     & 54.4 & 29.4 & 0.88$\times$ \\
Qwen3-VL-8B        & 57.4 & 28.1 & 0.84$\times$ \\
\midrule
\textit{Random}    & 25.0 & 33.3 & 1.00$\times$ \\
\bottomrule
\end{tabular}
\end{table}

\section{Per-Diagnostic Frame Sweep}
\label{app:frame_sweep}

\begin{figure*}[h]
  \centering
  \includegraphics[width=\linewidth]{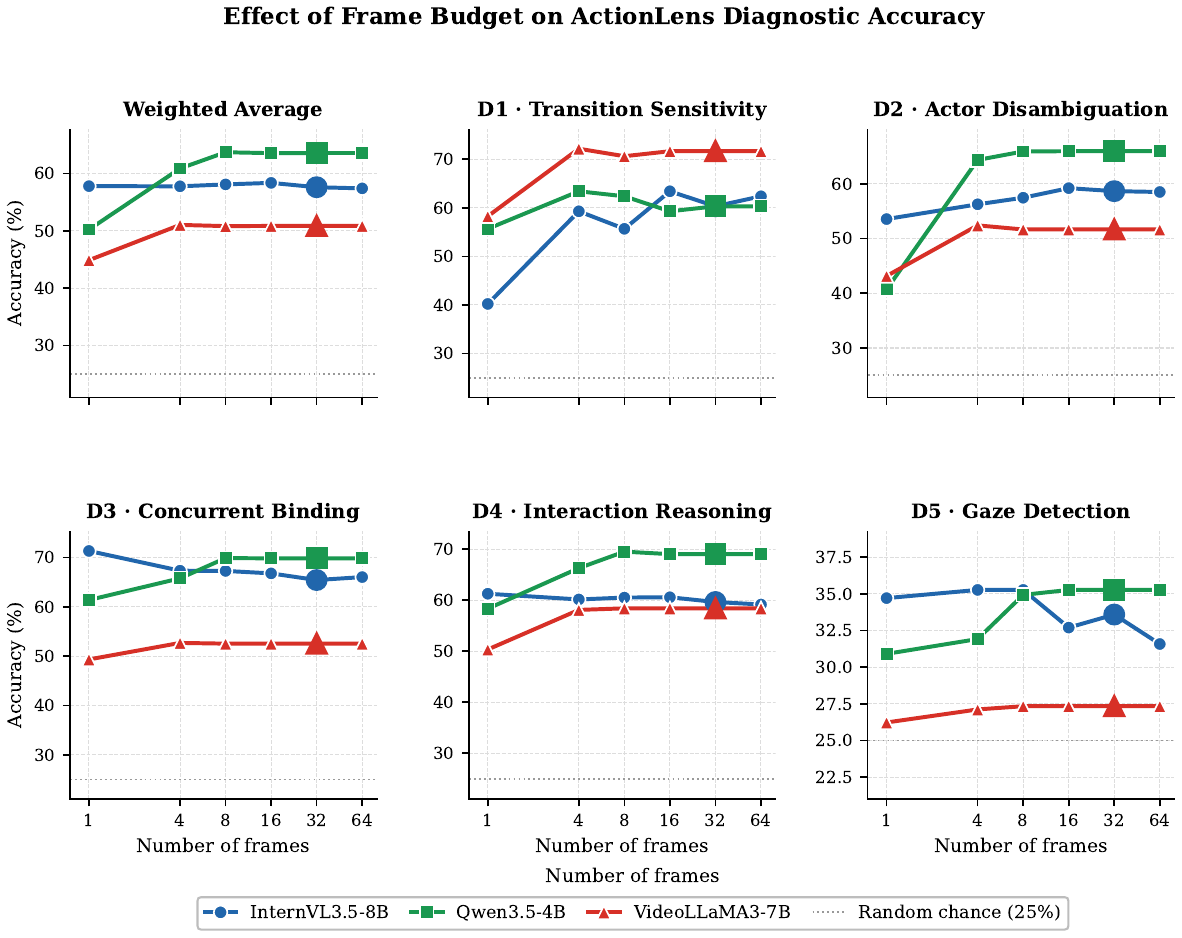}
  \caption{\textbf{Per-diagnostic frame sweep.}
    Transition Sensitivity is the most frame-hungry: all three models
    gain substantially from 1 to 8 frames on this diagnostic.
    Concurrent Binding peaks at 1 frame — it is a snapshot task.
    Actor Disambiguation and Interaction Reasoning show strong frame sensitivity for Qwen3.5-4B but near-flat
    curves for InternVL3.5-8B, suggesting different temporal aggregation
    mechanisms.}
  \label{fig:app_frame_diag}
\end{figure*}

Figure~\ref{fig:app_frame_diag} breaks down the frame-budget ablation by
diagnostic.  The weighted-average curves are in Figure~\ref{fig:frame_sweep}
in the main paper.

\begin{figure*}[h]
  \centering
  \includegraphics[width=\linewidth]{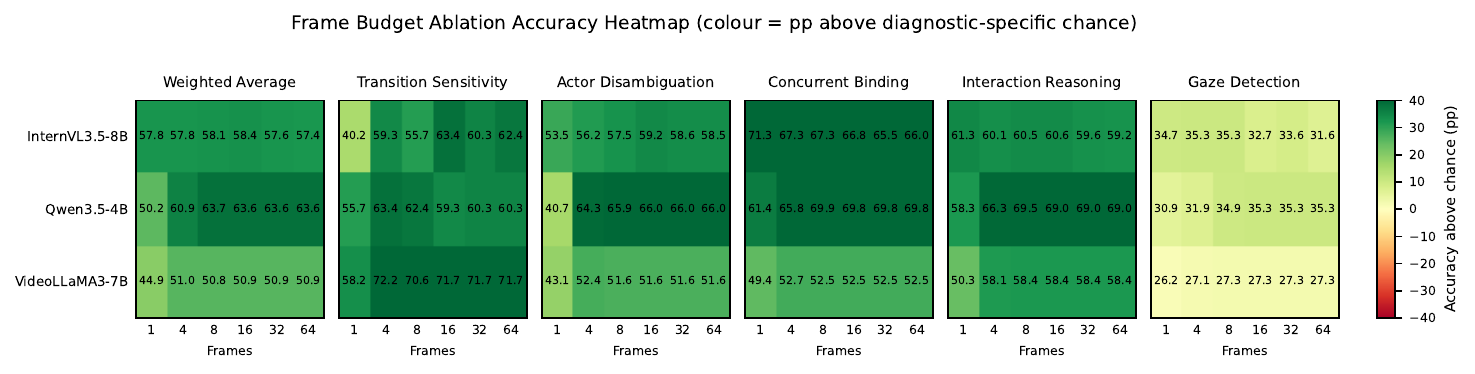}
  \caption{\textbf{Frame sweep heatmap.}
    Rows: (model, diagnostic) pairs. Columns: frame counts.
    Darker = higher accuracy.  Most cells stop changing after 8--16 frames;
    Transition Sensitivity is the primary exception.}
  \label{fig:app_frame_heatmap}
\end{figure*}

Figure~\ref{fig:app_frame_heatmap} shows the same data as a heatmap over
all (model, diagnostic, frame count) combinations, making the saturation
point easy to read off per cell.

\section{Text-Only Ablation: Full Results}
\label{app:text_only}

Five models answered the same questions with and without video. The no-video
condition replaced visual input with a blank context while preserving the
question and answer options. Table~\ref{tab:text_only} reports all five
diagnostics and the question-micro aggregate.

\begin{table}[t]
\centering
\small
\caption{%
  \textbf{Text-only ablation.}
  Accuracy (\%) on matched video and blank-visual conditions.
  $\Delta$ is text-only minus video; negative values favor video.
}
\label{tab:text_only}
\begin{tabular}{l l r r r r r r}
\toprule
& &
\makecell[c]{\textbf{Trans.}\\{\small Sens.}} &
\makecell[c]{\textbf{Actor}\\{\small Disambig.}} &
\makecell[c]{\textbf{Conc.}\\{\small Binding}} &
\makecell[c]{\textbf{Inter.}\\{\small Reasoning}} &
\makecell[c]{\textbf{Gaze}\\{\small Det.}} &
\makecell[c]{\textbf{W.}\\{\small Avg.}} \\
\midrule

\multirow{5}{*}{\rotatebox[origin=c]{90}{\scriptsize Video}}
& InternVL3.5-38B   & 68.0 & 65.5 & 68.4 & 68.7 & 38.6 & 63.6 \\
& LLava-Video-32B   & 60.3 & 45.0 & 42.0 & 51.0 & 28.7 & 43.8 \\
& Gemma-4-31B       & 63.9 & 63.0 & 72.9 & 61.6 & 42.0 & 62.8 \\
& Qwen3.5-9B        & 64.9 & 64.6 & 69.1 & 70.3 & 37.6 & 63.7 \\
& Qwen3-VL-8B       & 61.9 & 57.4 & 70.9 & 66.2 & 30.8 & 60.1 \\

\midrule

\multirow{5}{*}{\rotatebox[origin=c]{90}{\scriptsize Text only}}
& InternVL3.5-38B   & 69.6 & 43.3 & 40.9 & 43.1 & 23.9 & 40.7 \\
& LLava-Video-32B   & 49.0 & 31.0 & 19.3 & 37.1 & 24.4 & 28.6 \\
& Gemma-4-31B       & 27.3 & 23.4 & 28.7 & 28.8 & 46.0 & 29.4 \\
& Qwen3.5-9B        & 58.8 & 41.5 & 30.9 & 49.3 & 24.0 & 38.4 \\
& Qwen3-VL-8B       & 57.7 & 33.8 & 44.5 & 30.1 & 24.1 & 35.5 \\

\midrule

\multirow{5}{*}{\rotatebox[origin=c]{90}{\scriptsize $\Delta$}}
& InternVL3.5-38B   & \cellcolor{red!12}$+$1.6  & $-$22.2 & $-$27.5 & $-$25.6 & $-$14.7 & $-$22.9 \\
& LLava-Video-32B   & $-$11.3 & $-$14.0 & $-$22.7 & $-$13.9 & $-$4.3  & $-$15.2 \\
& Gemma-4-31B       & $-$36.6 & $-$39.6 & $-$44.2 & $-$32.8 & \cellcolor{red!12}$+$4.0  & $-$33.4 \\
& Qwen3.5-9B        & $-$6.1  & $-$23.1 & $-$38.2 & $-$21.0 & $-$13.6 & $-$25.3 \\
& Qwen3-VL-8B       & $-$4.2  & $-$23.6 & $-$26.4 & $-$36.1 & $-$6.7  & $-$24.6 \\

\bottomrule
\end{tabular}
\end{table}

\paragraph{Where wording helps.}
Transition questions reveal partial action-sequence priors: Qwen3.5-9B
loses only 6.1 points without video, Qwen3-VL-8B loses 4.2, and
InternVL3.5-38B gains 1.6. The ``before/after'' wording can suggest
familiar transitions, so this diagnostic should not be interpreted as a
pure measure of visual change detection.

\paragraph{Where video matters.}
Actor disambiguation, concurrent binding, and interaction reasoning all
fall when video is removed, for every tested model. This is the main
shortcut check: answer wording alone does not recover the person-action
associations. Gaze detection behaves differently. Video scores remain
modest, and text-only performance varies substantially by model, so the
control does not isolate a single cause of the gaze failure.

\section{Limitations}
\label{app:limitations}

\paragraph{Domain.}
\DatasetName{} is derived entirely from AVA~v2.2, which consists of Hollywood films.
Performance gaps may not transfer directly to other video domains (e.g., sports broadcasts, surveillance footage, ego-centric video), where action distributions, camera motion, and occlusion patterns differ substantially.

\paragraph{Dataset scale and Transition Sensitivity coverage.}
Transition Sensitivity (D1) contains only 194 items, constrained by AVA's annotation density for clear action-boundary transitions.
Conclusions about this diagnostic should be interpreted with lower statistical power than the other four diagnostics.

\paragraph{Human reference annotation.}
The 1,194-item pooled human reference was annotated by four non-author raters, each assigned a disjoint set of items.
Each item received one response, and the pooled exact-match accuracy is reported as the human reference.
The retained questions were iteratively filtered for visual answerability, so the 91.0\% reference is specific to this retained item set.
The iterative quality-engineering rounds were partly conducted by the authors, as direct involvement was necessary to identify and fix structural failures; this introduces a potential familiarity bias in the quality-review process that cannot be fully controlled.

\paragraph{Closed-source model coverage.}
GPT-5.2 and Gemini~3~Flash are evaluated only on the 1,194-item review subset due to API cost constraints, not the full 6,701-item benchmark.
Direct comparison between closed-source and open-source full-dataset numbers should account for this scope difference.

\paragraph{Statistical uncertainty.}
Deterministic decoding removes sampling variation from repeated generations,
but the 6,701 questions are concentrated in 64 source videos. Different
source-video samples can therefore change the measured scores and gaps.
We use paired video-cluster bootstrap intervals and video-subset analyses
(Appendix~\ref{app:statistical_uncertainty}); small model differences should
not be interpreted as stable rankings when their paired intervals cross zero.

\section{Statistical Uncertainty and Video-Level Robustness}
\label{app:statistical_uncertainty}

\paragraph{Paired cluster bootstrap.}
For the 18 open-weight models evaluated on the full benchmark, we draw 10,000
bootstrap samples of the 64 source videos with replacement. Each draw keeps
all questions from a sampled video together, and the same draws are used for
every model. We report percentile 95\% intervals for question-micro accuracy
(each question has equal weight) and diagnostic-macro accuracy (each of the five
diagnostics has equal weight). Paired model-difference intervals are calculated
within each common draw. The intervals characterize sensitivity to the sampled
videos, not variation from repeated deterministic decoding.

\begin{table}[h]
\centering\small
\setlength{\tabcolsep}{4pt}
\caption{\textbf{Full-set accuracy and video-cluster 95\% intervals (\%).}
All 18 models answer the same 6,701 questions from 64 source videos.
Micro weights questions equally; macro weights diagnostics equally.}
\label{tab:bootstrap_all_models}
\begin{tabular}{lcc}
\toprule
\textbf{Model} & \textbf{Question micro} & \textbf{Diagnostic macro} \\
\midrule
Qwen3.5-4B          & 63.6 [61.0, 66.1] & 60.1 [57.5, 62.5] \\
Qwen3.5-9B          & 63.7 [61.3, 66.1] & 61.3 [58.9, 63.5] \\
Qwen3.5-27B         & 68.8 [66.4, 71.2] & 65.6 [63.1, 68.0] \\
Qwen3-VL-4B         & 59.3 [56.8, 61.9] & 57.8 [55.4, 60.1] \\
Qwen3-VL-8B         & 60.1 [57.6, 62.7] & 57.4 [54.8, 60.0] \\
Qwen3-VL-32B        & 62.8 [60.0, 65.7] & 60.7 [57.5, 63.7] \\
Qwen2.5-VL-32B      & 59.3 [56.4, 62.3] & 58.0 [55.3, 60.7] \\
InternVL3.5-4B      & 54.7 [52.2, 57.2] & 53.4 [50.8, 55.7] \\
InternVL3.5-8B      & 57.6 [55.2, 60.0] & 55.5 [53.2, 57.9] \\
InternVL3.5-38B     & 63.6 [61.4, 65.9] & 61.8 [59.3, 64.2] \\
LLaVA-Video-7B      & 40.6 [38.8, 42.3] & 41.1 [39.3, 42.9] \\
LLaVA-Video-32B     & 43.8 [41.7, 45.9] & 45.4 [43.2, 47.6] \\
LLaVA-OneVision-7B  & 48.7 [46.0, 51.6] & 49.9 [47.2, 52.6] \\
VideoLLaMA3-7B      & 50.9 [48.8, 52.9] & 52.3 [50.2, 54.4] \\
Gemma-3-4B          & 27.7 [26.4, 29.2] & 31.0 [29.4, 32.6] \\
Gemma-3-27B         & 33.5 [31.8, 35.2] & 38.4 [36.6, 40.2] \\
Gemma-4-E4B         & 49.2 [46.5, 52.0] & 47.2 [44.9, 49.5] \\
Gemma-4-31B         & 62.8 [60.1, 65.6] & 60.7 [58.6, 62.8] \\
\bottomrule
\end{tabular}
\end{table}

\begin{table}[h]
\centering\scriptsize
\setlength{\tabcolsep}{3pt}
\caption{\textbf{Paired full-set differences in percentage points.}
Each cell is estimate [video-cluster 95\% interval], from the same 10,000
paired resamples. Positive values favor the first model.}
\label{tab:bootstrap_comparisons}
\begin{tabular}{lcc}
\toprule
\textbf{First model $-$ second model} & \textbf{Question micro} & \textbf{Diagnostic macro} \\
\midrule
Qwen3.5-27B $-$ Qwen3.5-4B         & $+$5.18 [3.55, 6.68]   & $+$5.52 [3.32, 7.73] \\
Qwen3.5-27B $-$ Qwen3.5-9B         & $+$5.06 [3.58, 6.59]   & $+$4.29 [2.20, 6.50] \\
InternVL3.5-38B $-$ InternVL3.5-4B & $+$8.85 [7.00, 10.61] & $+$8.45 [6.69, 10.15] \\
InternVL3.5-38B $-$ InternVL3.5-8B & $+$5.97 [4.32, 7.60]  & $+$6.30 [4.44, 8.17] \\
LLaVA-Video-32B $-$ LLaVA-Video-7B & $+$3.22 [0.81, 5.65] & $+$4.23 [1.79, 6.67] \\
Qwen3.5-27B $-$ InternVL3.5-38B   & $+$5.18 [3.61, 6.75]   & $+$3.76 [1.69, 5.96] \\
InternVL3.5-38B $-$ Qwen3.5-4B    & 0.00 [$-$1.74, 1.59]  & $+$1.75 [$-$0.85, 4.36] \\
Qwen3.5-27B $-$ LLaVA-Video-32B   & $+$24.98 [22.61, 27.35] & $+$20.22 [17.64, 22.66] \\
InternVL3.5-38B $-$ LLaVA-Video-32B & $+$19.80 [17.71, 21.93] & $+$16.45 [13.84, 18.99] \\
Qwen3.5-27B $-$ Qwen3-VL-32B      & $+$5.92 [4.55, 7.34] & $+$4.89 [2.89, 6.95] \\
Qwen3.5-27B $-$ Gemma-4-31B       & $+$5.94 [4.10, 7.70] & $+$4.92 [2.78, 6.89] \\
Qwen3-VL-32B $-$ Gemma-4-31B      & $+$0.01 [$-$1.90, 1.88] & $+$0.04 [$-$2.35, 2.35] \\
Gemma-4-E4B $-$ Gemma-3-27B       & $+$15.73 [12.67, 18.81] & $+$8.77 [6.13, 11.54] \\
\bottomrule
\end{tabular}
\end{table}

\paragraph{Source-video concentration and alternative weighting.}
The 64 sources contribute 36--199 questions each (median 100); the ten largest
sources contribute 25.6\% of all questions, and 52 sources appear in every
diagnostic. Video-cluster 95\% interval half-widths are 1.29--2.53 times
the corresponding independent-question intervals. Equal-video weighting
(one mean accuracy per source video) changes each model's question-micro
score by at most 1.58 points. Under equal-video weighting, Qwen3.5-27B
leads InternVL3.5-38B by 4.76 points [3.02, 6.48], and
LLaVA-Video-32B by 25.10 points [22.72, 27.48].

\begin{table}[h]
\centering\small
\setlength{\tabcolsep}{4pt}
\caption{\textbf{Ranking stability under distinct-video subsampling.}
Each row uses 10,000 subsets of videos without replacement and
diagnostic-macro scoring; intervals summarize the subset distribution.}
\label{tab:video_subset_stability}
\begin{tabular}{rcc}
\toprule
\textbf{Videos retained} & \textbf{Mean Spearman $\rho$ [95\% interval]} & \textbf{Leader retained} \\
\midrule
8  & 0.946 [0.878, 0.986] & 79.5\% \\
16 & 0.971 [0.926, 0.994] & 96.6\% \\
32 & 0.987 [0.963, 0.998] & 100.0\% \\
48 & 0.993 [0.979, 1.000] & 100.0\% \\
\bottomrule
\end{tabular}
\end{table}

\paragraph{Reducing within-video correlation.}
Across all 64 leave-one-video-out folds, the minimum rank correlation with
the full diagnostic-macro ranking is 0.996, and no single omission changes
any model's score by more than 0.54 points. Sampling one question per
populated video--diagnostic cell (303 cells) still yields mean rank
correlation 0.941 [0.882, 0.981] across 10,000 repetitions. These checks
show that the broad model-family pattern is not driven by a few prolific
source videos; narrow diagnostic leaderboards remain less stable, especially
when very few videos are retained.

\section{Portability to Sports Video}
\label{app:portability}

MultiSports~\cite{li2021multisports} supplies person action tubes in sports
broadcasts rather than AVA's movie annotations. We mapped its official
validation annotations to the actor-disambiguation template, selected two
visible people with distinct actions, and rendered tracked target boxes and
wrong-actor distractors. The automatic pool contains 500 candidate questions
from 120 volleyball videos; all three models saw the same items.

\begin{table}[h]
\centering\small
\setlength{\tabcolsep}{5pt}
\caption{\textbf{MultiSports actor-disambiguation pilot.} Accuracy and
wrong-actor selections are measured on 500 automatically generated
candidates; intervals resample the 120 source videos. Wrong-actor rate is
conditional on an incorrect answer.}
\label{tab:multisports_pilot}
\begin{tabular}{lcc}
\toprule
\textbf{Model} & \textbf{Accuracy [95\% CI]} & \textbf{Wrong actor / errors} \\
\midrule
Qwen3.5-9B      & 57.4 [53.0, 61.7] & 51/213 (23.9\%) \\
InternVL3.5-8B  & 57.6 [52.0, 63.2] & 34/212 (16.0\%) \\
Gemma-4-E4B     & 48.0 [43.5, 52.6] & 46/260 (17.7\%) \\
\bottomrule
\end{tabular}
\end{table}

Nominal wrong-actor selections occur in all three models, but their share of errors
is below the 33.3\% expected from uniformly choosing among the three wrong
options. This suggests the error can recur outside AVA, but does not show
enrichment to the same degree. These candidates have not passed
ActionLens's visual audit and failure-coded validation; their scores are
directional evidence about portability, not a second finalized benchmark.

\paragraph{Minimum evidence for a new domain.}
The construction recipe needs (i) continuous target identity through the
clip, (ii) unique action or role support for the queried person and time,
and (iii) a visual audit that the rendered reference and choices admit one
defensible answer. Dense per-second labels make these checks easier, but
they are not inherently required. Actor disambiguation and concurrent
binding can start from verified tracks and distinct actions; transition
sensitivity needs action evidence on both sides of a time boundary;
interaction reasoning needs a directed relation between identifiable
people; gaze detection needs a visually supportable gaze label. With weaker
source labels, human adjudication becomes part of ground-truth creation.

\paragraph{Adapter guide.}
To port the pipeline: map source tracks and labels into a common evidence
schema; select only diagnostics supported by that evidence; tune filtering
thresholds to the domain; run a small failure-coded pilot and repair global
generation rules; then freeze and independently validate the resulting
questions. New-domain scores should be reported separately until item
validity is established for each source.

\section{Prompting Sensitivity}
\label{app:prompting}

We compared direct answer selection with zero-shot step-by-step prompting
\cite{kojima2022zeroshot} and a plan-then-solve prompt
\cite{wang2023planandsolve} on the same 1,194-item review subset. Direct
answers used a 16-token limit; the two reasoning prompts allowed 512 tokens
and requested a final option letter. Every output yielded a valid extracted
answer. These are prompt variants inspired by the cited methods, not exact
replications of their original text-only experiments.

\begin{table}[h]
\centering\scriptsize
\setlength{\tabcolsep}{3pt}
\caption{\textbf{Prompting sensitivity (\%).} All rows use the same
1,194 questions; columns name the five diagnostics. Overall weights
questions equally.}
\label{tab:prompting_sensitivity}
\begin{tabular}{llrrrrrr}
\toprule
\textbf{Model} & \textbf{Prompt} & \textbf{Trans.} & \textbf{Actor} & \textbf{Concurrent} & \textbf{Interaction} & \textbf{Gaze} & \textbf{Overall} \\
\midrule
Qwen3.5-4B & Direct & 60.3 & 67.2 & 74.0 & 66.8 & 33.6 & 60.4 \\
             & Step-by-step & 64.9 & 70.0 & 65.2 & 51.6 & 40.8 & 58.2 \\
             & Plan-and-solve & 58.8 & 64.0 & 68.0 & 49.6 & 42.4 & 56.4 \\
Qwen3.5-27B & Direct & 64.9 & 70.0 & 78.4 & 76.4 & 39.6 & 65.9 \\
              & Step-by-step & 64.9 & 67.6 & 77.2 & 60.8 & 44.4 & 62.9 \\
              & Plan-and-solve & 58.8 & 64.4 & 72.8 & 39.6 & 46.0 & 56.2 \\
InternVL3.5-4B & Direct & 61.9 & 60.4 & 68.4 & 46.8 & 23.6 & 51.8 \\
                & Step-by-step & 66.0 & 59.6 & 67.6 & 50.0 & 25.2 & 53.1 \\
                & Plan-and-solve & 70.1 & 52.4 & 64.8 & 50.0 & 34.0 & 53.5 \\
\bottomrule
\end{tabular}
\end{table}

Reasoning prompts improve gaze detection for these models and modestly
improve InternVL3.5-4B overall, but lower overall accuracy for both Qwen3.5
models. The strongest observed decline is on interaction reasoning:
Qwen3.5-27B falls from 76.4\% with direct answers to 39.6\% with
plan-and-solve. Prompting is therefore consequential, but neither tested
reasoning prompt closes the gap to the 91.0\% pooled human reference.

\section{Target-Box Color Swap}
\label{app:color_swap}

Rendered marks can change visual-model behavior
\cite{shtedritski2023redcircle}. We swapped red and blue on all 2,000
actor-disambiguation items while querying the \emph{same physical person};
the clip interval, options, answer letter, and option order stayed fixed.
Positive differences below favor a red target box. Intervals use 10,000
paired bootstrap resamples of the 63 source videos.

\begin{table}[h]
\centering\small
\setlength{\tabcolsep}{4pt}
\caption{\textbf{Same-target red/blue swap.} Accuracy (\%) and paired
red-minus-blue difference in percentage points.}
\label{tab:color_swap}
\begin{tabular}{lccc}
\toprule
\textbf{Model} & \textbf{Blue} & \textbf{Red} & \textbf{Difference [95\% CI]} \\
\midrule
Qwen3.5-27B     & 67.70 & 67.25 & $-$0.45 [$-$1.58, 0.65] \\
InternVL3.5-8B   & 58.65 & 58.65 & 0.00 [$-$1.50, 1.50] \\
Gemma-4-E4B      & 47.90 & 50.00 & $+$2.10 [0.30, 4.04] \\
VideoLLaMA3-7B   & 51.65 & 49.70 & $-$1.95 [$-$5.12, 0.99] \\
\bottomrule
\end{tabular}
\end{table}

There is no consistent aggregate preference for red or blue across models.
Gemma-4-E4B favors red in this experiment, while the other clustered
intervals include zero. Individual decisions can still flip: predictions
agree between color conditions on 77.5--92.3\% of items, depending on model.
The wrong-actor distractor remains a frequent choice under both colors.

\section{Scope of the Human-Review Subset}
\label{app:subset_scope}

The 1,194-item subset contains every transition-sensitivity question and
250 fixed-seed, round-robin samples from each other diagnostic. It covers
all 64 source videos in aggregate. We compare six open models evaluated
under the same protocol on both scopes, keeping question-micro and
diagnostic-macro aggregation separate.

\begin{table}[h]
\centering\small
\setlength{\tabcolsep}{4pt}
\caption{\textbf{Full-set versus human-review subset accuracy (\%).}
Micro weights questions equally; macro weights the five diagnostics equally.
The subset changes diagnostic proportions, so these aggregates need not
shift together.}
\label{tab:subset_scope}
\begin{tabular}{lcccc}
\toprule
& \multicolumn{2}{c}{\textbf{Question micro}} & \multicolumn{2}{c}{\textbf{Diagnostic macro}} \\
\textbf{Model} & \textbf{Full} & \textbf{Subset} & \textbf{Full} & \textbf{Subset} \\
\midrule
Qwen3.5-27B      & 68.8 & 65.9 & 65.6 & 65.9 \\
InternVL3.5-38B   & 63.6 & 61.3 & 61.8 & 61.6 \\
Gemma-4-31B      & 62.8 & 62.3 & 60.7 & 62.4 \\
Qwen3.5-4B       & 63.6 & 60.4 & 60.1 & 60.4 \\
InternVL3.5-4B    & 54.7 & 51.8 & 53.4 & 52.2 \\
LLaVA-Video-32B  & 43.8 & 44.1 & 45.4 & 44.9 \\
\bottomrule
\end{tabular}
\end{table}

The full-set leader remains the subset leader. Across these six models,
the mean absolute full/subset difference is 0.70 points for diagnostic
macro and 2.03 for question micro; the largest question-micro shift is
3.20 points. Thus the subset supports broad comparisons on matched items,
but its absolute accuracy is not interchangeable with the full set's.

\section{Reproducibility}
\label{app:repro}

\paragraph{Code and data.}
Full pipeline, task configs, and raw evaluation logs are at
\url{https://anonymous.4open.science/r/lmms-eval-2276}.

\paragraph{One-command evaluation.}
\begin{quote}\small\tt
bash examples/eval\_actionlens.sh
\end{quote}
Videos stream automatically; no manual data preparation is needed.
Set environment variables at the top of the script to override model, tasks,
device, and output directory.

\paragraph{Determinism.}
All runs use \texttt{temperature=0}, \texttt{do\_sample=False},
\texttt{num\_beams=1}.  Results are fully deterministic given identical model
weights and hardware.

\paragraph{Pooled human reference protocol.}
Annotations were collected by four non-author raters on the 1,194-item review subset, with disjoint item assignments.
Each item received one independent response without access to model predictions.
Raters watched each clip in full before answering; no time limit was set.
The resulting 1,086/1,194 accuracy (91.0\%) is the pooled human reference.
Note: the iterative quality-engineering rounds (Section~\ref{app:quality}) were partly conducted by the authors, as direct involvement was necessary to diagnose and resolve structural dataset failures.

\end{document}